\documentclass[10pt,journal,compsoc]{IEEEtran}

\usepackage{times}
\usepackage{epsfig}
\usepackage{graphicx}
\usepackage{amsmath}
\usepackage{amssymb}
\usepackage{color}
\usepackage{algorithm}
\usepackage{algorithmic}
\usepackage{subfigure}
\usepackage{epstopdf}
\usepackage{booktabs}
\usepackage{array}
\usepackage{multirow}
\usepackage{hhline}
\usepackage{makecell}
\usepackage{framed}
\usepackage{enumitem}

\newcommand{\Eref}[1]{Equation~(\ref{#1})}
\newcommand{\Fref}[1]{Figure~\ref{#1}}

\newcommand{\fref}[1]{Fig.~\ref{#1}}

\graphicspath{{images/}{images/restoration/}}

\begin{document}

\title{ A General Decoupled Learning Framework  for Parameterized Image Operators }

\author{Qingnan Fan, Dongdong Chen,~\IEEEmembership{Member,~IEEE},
        Lu Yuan, Gang Hua,~\IEEEmembership{Fellow,~IEEE}, Nenghai Yu
        and Baoquan Chen

\IEEEcompsocitemizethanks{\IEEEcompsocthanksitem Qingnan Fan is with Computer Science Department, Stanford University, Stanford, California 94305, US.\protect\\
E-mail: fqnchina@gmail.com
\IEEEcompsocthanksitem Dongdong Chen and Nenghai Yu are with Department of Electronic Engineering and Information Science, University of Science and Technology of China, Hefei, Anhui 230026, China.\protect\\
E-mail: cd722522@mail.ustc.edu.cn, ynh@ustc.edu.cn
\IEEEcompsocthanksitem Lu Yuan is with Microsoft Research, Redmond, Washington 98052, USA.\protect\\
E-mail: \{luyuan\}@microsoft.com
\IEEEcompsocthanksitem Gang Hua is with Wormpex AI Research, Bellevue, Washington 98004, USA.\protect\\
E-mail: \{ganghua\}@gmail.com
\IEEEcompsocthanksitem Baoquan Chen is with Peking University, Beijing 100871, China.\protect\\
E-mail: baoquan@pku.edu.cn
}
\thanks{Manuscript received September 27, 2018; revised April 17, 2019; accepted June 25, 2019. This work was supported in part by: National 973 Program (2015CB352501), NSFC-ISF (61561146397), the Natural Science Foundation of China under Grant U1636201 and 61629301. (Qingnan Fan and Dongdong Chen are co-first authors.)}
}

\markboth{Journal of \LaTeX\ Class Files, August~2019}%
{Shell \MakeLowercase{\textit{et al.}}: Bare Demo of IEEEtran.cls for Computer Society Journals}

\IEEEtitleabstractindextext{%
\begin{abstract}
Many different deep networks have been used to approximate, accelerate or improve traditional image operators. Among these traditional operators, many contain parameters which need to be tweaked to obtain the satisfactory results, which we refer to as ``parameterized image operators''. However, most existing deep networks trained for these operators are only designed for one specific parameter configuration, which does not meet the needs of real scenarios that usually require flexible parameters settings. To overcome this limitation, we propose a new decoupled learning algorithm to learn from the operator parameters to dynamically adjust the weights of a deep network for image operators, denoted as the \textit{base} network. The learned algorithm is formed as another network, namely the \textit{weight learning} network, which can be end-to-end jointly trained with the \textit{base} network.  Experiments demonstrate that the proposed framework can be successfully applied to many traditional parameterized image operators. To accelerate the parameter tuning for practical scenarios, the proposed framework can be further extended to dynamically change the weights of only one single layer of the \emph{base} network while sharing most computation cost. We demonstrate that this cheap parameter-tuning extension of the proposed decoupled learning framework even outperforms the state-of-the-art alternative approaches.
\end{abstract}
}

\maketitle

\IEEEdisplaynontitleabstractindextext

\IEEEpeerreviewmaketitle

\IEEEraisesectionheading{
\section{Introduction}\label{sec:introduction}
}

\IEEEPARstart{I}{mage} operators are fundamental building blocks for many computer vision tasks, such as image filtering and restoration. To obtain the desired results, many of these operators contain some parameters that need to be tweaked. We refer them as ``parameterized image operators'' in this paper. For example, parameters controlling the smoothness strength are widespread in most smoothing methods, and a parameter denoting the target upsampling scalar is always used in image super resolution.

Recently, many CNN based methods \cite{fan2017generic,kim2016accurate,xu2015deep} have been proposed to approximate, accelerate or improve these parameterized image operators  and achieved significant progress. However, we observe that the networks in these methods are often only trained for one specific parameter configuration, such as edge-preserving filtering \cite{fan2017generic} with a fixed smoothness strength, or super resolving low-quality images \cite{kim2016accurate} with a particular downsampling scale. Many different models need to be retrained for different parameter settings, which is both storage-consuming and time-consuming. It also prohibits these deep learning solutions from being applicable and extendable to a much broader corpus of images.

In fact, given a specific network structure, when training separated networks for different parameter configurations $\overrightarrow{\gamma}_k$ as \cite{fan2017generic, kim2016accurate, xu2015deep}, the learned weights $W_k$ are highly unconstrained and probably very different for each $\overrightarrow{\gamma}_k$. But can we find a common convolution weight space for different configurations by explicitly building their relationships? Namely, $W_k = h(\overrightarrow{\gamma}_k)$, where $h$ can be a linear or non-linear function. In this way, we can adaptively change the weights of the single target network based on $h$ in the runtime, thus enabling continuous parameter control.

To verify our hypothesis, we propose the first decoupled learning framework for parameterized image operators by decoupling the weights from the target network structure. Specifically, we employ a simple \textit{weight learning} network $\mathcal{N}_{weight}$ as $h$ to directly learn the convolution weights of one task-oriented \textit{base} network $\mathcal{N}_{base}$.
These two networks can be trained end-to-end. During the runtime, the \textit{weight learning} network will dynamically update the weights of the \textit{base} network according to different input parameters, thus making the \textit{base} network generate different objective results. This should be a very useful feature in scenarios where users want to adjust and select the most visually pleasant results interactively.

We demonstrate the effectiveness of the proposed framework for many different types of applications, such as edge-preserving image filtering with different degrees of smoothness, image super resolution with different scales of blurring, and image denoising with different magnitudes of noise. We also demonstrate the extensibility of our proposed framework on multiple input parameters for a specific application, and combination of multiple different image processing tasks. Experimental results demonstrate that the proposed framework is able to achieve almost as good results as the one solely trained for a single parameter value.

Despite of its generality and flexibility, all the convolution weights of $\mathcal{N}_{base}$ need to be updated during the runtime. In the other words, the whole network needs to be re-evaluated when a new parameter is selected. This is very time-consuming and unacceptable in real user scenarios. To be adaptive for practical applications, the proposed framework can be further extended to dynamically change the weights of only one single layer while sharing most of the computation. With our default $\mathcal{N}_{base}$ network structure, experiments demonstrate this cheap parameter-tuning version outperforms the state-of-the-art methods \cite{chen2016bilateral,gharbi2017deep} by a large margin.

As an extra bonus, the proposed framework makes it easy to analyze the underlying working principle of the trained task-oriented network by visualizing different parameters. The knowledge gained from this analysis may inspire more promising research in this area. To sum up, the contributions of this paper lie in the following four aspects.

\begin{itemize}
\item We propose the first decoupled learning framework for parameterized image operators, where a \textit{weight learning} network is learned to adaptively predict the weights for the task-oriented \textit{base} network in the runtime.
\item We show that the proposed framework can be learned to incorporate many different parameterized image operators and achieve very competitive performance with the one trained for a single specific parameter or operator.
\item We extend our framework to enable cheap parameter tuning for real user scenarios, which outperforms many state-of-the-art methods by a large margin.
\item We provide a unique perspective to understand the working principle of the trained task-oriented network with some valuable analysis and discussion, which may inspire more promising research in this area.
\end{itemize}

\section{Related Work}
In the past decades, many different image operators have been proposed for low level vision tasks. Previous work~\cite{xu2011image,karacan2013structure,xu2012structure,zhang2014rolling} proposed different priors to smooth images while preserving salient structures. Some papers~\cite{buades2005non,elad2006image} utilized the spatial relationship and redundancy to remove unpleasant noise in the image. Other papers~\cite{yang2010image,sun2008image,tipping2003bayesian} aimed to recover a high-resolution image from a low-resolution image. Among them, many operators are allowed to tune some built-in parameters to obtain different results, which is the focus of this paper.

With the development of deep learning techniques, many different neural networks are proposed to approximate, accelerate and improve these operators \cite{fan2017generic,xu2015deep,dong2016image,jain2009natural,liu2016learning}.
But their common limitation is that one model can only handle one specific parameter setting. To enable all other parameters, enormous different models need to be retrained, which is both storage-consuming and time-consuming.
By contrast, our proposed framework allows us to input continuous parameters to dynamically adjust the weights of the task-oriented \textit{base} network. Moreover, it can even be applied to multiple different parameterized operators with one single network.

Recently, Chen {\em et al.}~\cite{chen2017fast} conducted a naive extension for parameterized image operators by concatenating the parameters as extra input channels to the network. Compared to their method, where both the network structure and weights maintain the same for different  parameters, the weights of our \textit{base} network are adaptively changed. Experimentally we find our framework outperforms their strategy by integrating multiple image operators. By decoupling the network structure and weights, our proposed framework also makes it easier to analyze the underlying working principle of the trained task-oriented network, rather than leaving it as a black box as in many previous works like ~\cite{chen2017fast}.

To enable practical image processing on mobile devices, a simple scheme to accelerate an operator is to apply it at a low-resolution image then upsample the result by reintroducing high-resolution details, which is used in bilateral upsampling \cite{kopf2007joint} and the fast guided filter \cite{he1505fast}. To unify and generalize these two methods, \cite{chen2016bilateral} presents bilateral guided upsampling, which is further extended to \cite{gharbi2017deep} by incorporating such a technique into an end-to-end trained deep network. Though \cite{chen2016bilateral} and \cite{gharbi2017deep} are able to run very efficiently on even CPU devices, their resultant image quality is not good enough. By contrast, our cheap parameter-tuning extension directly runs the approximated operators on the original image resolution by sharing most of the computation costs for different operators, which not only enables real-time parameter adjustment on CPU devices but is also able to demonstrate its superior performance over these recent state-of-the-art approaches \cite{chen2016bilateral,gharbi2017deep} by a large margin.

Our method is also related to evolutionary computing and meta learning. Schmidhuber~\cite{schmidhuber1992learning} suggested the concept of fast weights in which one network can produce context-dependent weight changes for a second network. Some other works~\cite{Andrychowicz2016,Wichrowska2017a,Chen2017a} casted the design of an optimization algorithm as a learning problem, Recently, Ha {\em et al.}~\cite{ha2016hypernetworks} proposed to use a static hypernetwork to generate weights for a convolutional neural network on MNIST and Cifar classification. They also leverage a dynamic hypernetwork to generate weights of recurrent networks for a variety of sequence modelling tasks. The purpose of their paper is to exploit weight sharing property across different convolution layers. But in our cases, we pay more attention to the common shared property among numerous input parameters and many different image operators.

\section{Method}

\subsection{Problem Definition and Motivation}
The input color image and the target parameterized image operators are denoted as $\mathcal{I}$ and $f(\overrightarrow{\gamma}, \mathcal{I})$ respectively. $f(\overrightarrow{\gamma}, \mathcal{I})$  transforms the content of $\mathcal{I}$ locally or globally without changing its dimension. $\overrightarrow{\gamma}$ denotes the parameters which determine the transform degree of $f$ and may be a single value or a multi-value vector. For example, in $L_0$ smoothing\cite{l0smoothing2011}, $\overrightarrow{\gamma}$ is the balance weight controlling the smoothness strength, while in RTV filter \cite{xu2012structure}, it includes one more spatial gaussian variance. In most cases, $f$ is a highly nonlinear process and solved by iterative optimization methods, which is very slow in runtime.

\begin{figure*}[t]
	\includegraphics[width=1.0\linewidth]{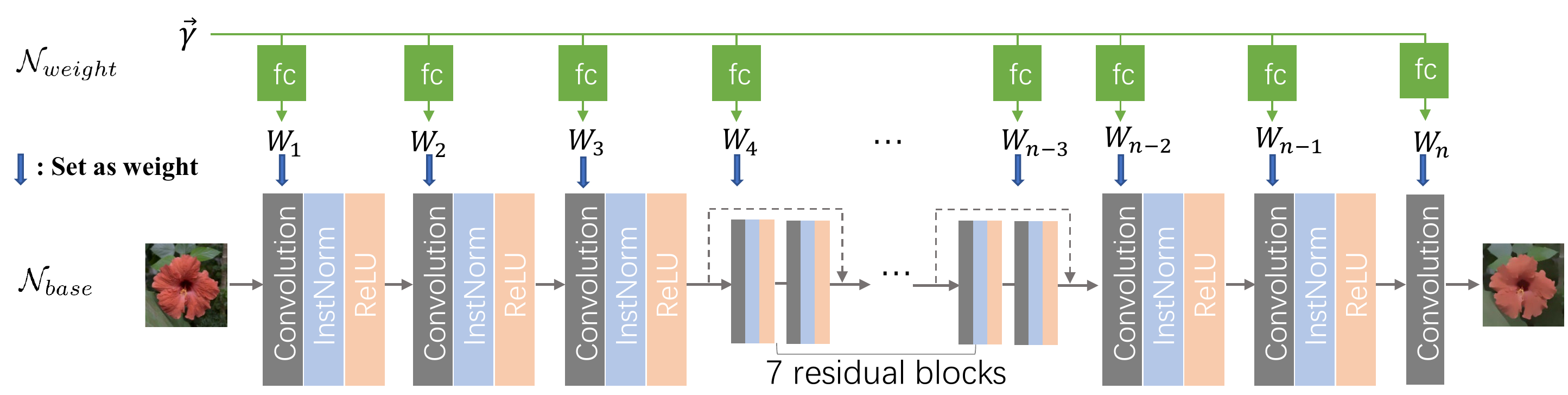}
	\caption{Our system consists of two networks: the above \textit{weight learning} network $\mathcal{N}_{weight}$ is designed to learn the convolution weights for the bottom \textit{base} network $\mathcal{N}_{base}$. Given a parameterized image operator constraint by $\protect\overrightarrow{\gamma}$,  these two networks are jointly trained, and $\mathcal{N}_{weight}$ will dynamically update the weights of $\mathcal{N}_{base}$ for different $\protect\overrightarrow{\gamma}$ in the inference stage.}
	\label{fg:arch}
\end{figure*}

Our goal is to implement parameterized operator $f$ with a base convolution network $\mathcal{N}_{base}$.  In previous methods like \cite{xu2015deep,liu2016learning}, given a specific network structure of $\mathcal{N}_{base}$, separated networks are trained for different parameter configurations $\overrightarrow{\gamma}_k$. In this way, the learned weights $\overrightarrow{W}_k$ of these separated networks are highly unconstrained and probably very different. But intuitively, for one specific image operator, the weights $\overrightarrow{W}_k$ of different $\overrightarrow{\gamma}_k$ might be related. So retraining separated models is too redundant. Motivated by this, we try to find a common weight space for different $\overrightarrow{\gamma}_k$ by adding a mapping constraint:  $\overrightarrow{W}_k = h(\overrightarrow{\gamma}_k)$, where $h$ can be a linear or non-linear function.

In this paper, we directly learn $h$ with another \textit{weight learning} network $\mathcal{N}_{weight}$ rather than design it by handcraft. Assuming  $\mathcal{N}_{base}$ is a fully convolutional network containing a total of $n$ convolution layers, we denote their weights as $\overrightarrow{W}_k=(W_1, W_2, ..., W_n)$ respectively, then

\begin{equation}
\begin{aligned}
(W_1, W_2, ..., W_n) = \mathcal{N}_{weight}(\overrightarrow{\gamma})
\end{aligned}
\end{equation}

where the input of $\mathcal{N}_{weight}$ is $\overrightarrow{\gamma}$ and the outputs are these weight matrices. In the training stage, $\mathcal{N}_{base}$ and $\mathcal{N}_{weight}$ can be jointly trained. In the inference stage, given different input parameter $\overrightarrow{\gamma}$, $\mathcal{N}_{weight}$ will adaptively change the weights of the target base network $\mathcal{N}_{base}$, thus enabling continuous parameter control.

Besides the original input image $\mathcal{I}$, the computed edge maps are shown to be a very important input signal for the target \emph{base} network in \cite{fan2017generic}. Therefore, we also pre-calculate the edge map $E$ of $\mathcal{I}$ and concatenate it to the original image as an extra input channel:
\begin{equation}
\begin{aligned}
E_{x,y} = \frac{1}{4}\sum_c(|\mathcal{I}_{x,y,c}-\mathcal{I}_{x-1,y,c}| + |\mathcal{I}_{x,y,c}-\mathcal{I}_{x+1,y,c}| \\ + |\mathcal{I}_{x,y,c}-\mathcal{I}_{x,y-1,c}| + |\mathcal{I}_{x,y,c}-\mathcal{I}_{x,y+1,c}|)
\end{aligned}
\end{equation}
where $x,y$ are the pixel coordinates and $c$ refers to the color channels.

To jointly train $\mathcal{N}_{base}$ and $\mathcal{N}_{weight}$, we simply use pixel-wise L2 loss in the RGB color space as \cite{chen2017fast} by default:
\begin{equation}
\begin{aligned}
\mathcal{L} = \lVert \mathcal{N}_{base}(\mathcal{N}_{weight}(\overrightarrow{\gamma}),\mathcal{I}, E) - f(\overrightarrow{\gamma}, \mathcal{I}) \rVert^2
\end{aligned}
\end{equation}

\subsection{Network Structure}\label{sec:basic}
As shown in \fref{fg:arch}, our \textit{base} network $\mathcal{N}_{base}$ follows a similar network structure as \cite{fan2017generic}. We employ 20 convolutional layers with the same $3\times3$ kernel size, among which the intermediate 14 layers are formed as residual blocks. Except the last convolution layer, all the former convolutional layers are followed by an instance normalization \cite{ulyanov2017improved} layer and a ReLU layer. To enlarge the receptive field of $\mathcal{N}_{base}$, the third convolution layer downsamples the dimension of feature maps by 1/2 using stride 2, and the third-to-last deconvolution layer (kernel size of $4\times4$) upsamples the downsampled feature maps to the original resolution symmetrically. In this way, the receptive field is effectively enlarged without losing too much image detail, and meanwhile the computation cost of intermediate layers is reduced. To further increase the receptive field, we also adopt the dilated convolution \cite{yu2015multi} as \cite{chen2017fast}, more detailed network structure can be found in the supplementary material.

In this paper, the \textit{weight learning} network $\mathcal{N}_{weight}$ simply consists of 20 fully connected (fc) layers by default. The $i_{th}$ fc layer is responsible to learn the weights $W_i$ for the $i_{th}$ convolutional layer, which can be written as following:
\begin{equation}
\begin{aligned}
W_i = A_i\overrightarrow{\gamma} + B_i, \qquad\forall i \in \{1,2,...,20\}
\end{aligned}
\label{eq:fc_weight}
\end{equation}
Where $A_i, B_i$ are the weight and bias of the $i_{th}$ fc layer. Assuming the parameter $\overrightarrow{\gamma}$ has a dimension of $m$ and $W_i$ has a dimension of $n_{wi}$. The dimension of $A_i$ and $B_i$ would be $n_{wi}\times m$ and $n_{wi}$ respectively.

Note in this paper, we don't intend to design an optimal network structure neither for the \textit{base} network $\mathcal{N}_{base}$ nor the \textit{weight learning} network $\mathcal{N}_{weight}$. On the contrary, we care more about whether it is feasible to learn the relationship between the weights of $\mathcal{N}_{base}$ and different parameter configurations $\overrightarrow{\gamma}$ even by such a simple \textit{weight learning} network $\mathcal{N}_{weight}$.

\subsection{Adaption to Cheap Parameter-Tuning}

In the above default setting, the weights of all the convolution layers in $\mathcal{N}_{base}$ are learned by the \emph{weight learning} network $\mathcal{N}_{weight}$ and would be dynamically changed with the input parameters. That is to say, if users want to do some parameter tuning for different visual effects, the whole network needs to be re-evaluated. However in order to obtain the best image quality, most current methods like \cite{fan2017generic,liu2016learning,chen2017fast}, including our default design, often requires to run through a complex deep networks and is very expensive for computational cost.

To tackle the efficiency issue, we further extend our framework from learning all the convolution weights to learning the weights of only one single layer. Since all the following layers behind this adjustable layer need to be re-evaluated when fed with varying input features, it is better to put this layer as deeper as possible in the \textit{base} network. In this way, the computation of all the preceding layers can be shared by different parameters, and only the layers after this single adjustable layer need to be re-evaluated. This is of great practicability for many real scenarios.

Specifically, that is to say,  only the weights $W_i$ of the $i$th layer is the function of input parameter $\overrightarrow{\gamma}$  as \Eref{eq:fc_weight} while all the remaining weights $W_k (k \neq i)$ are shared. During the runtime for parameter tuning, we only need to re-run the layers $k (k \geqslant i)$. Note that in our default network structure, $W_i$ can be either the weights of the convolutional layer or the scale and shift paramters in the following normalization layer. In this paper, by default, we choose the last instance normalization layer as the target layer and learn its scale and shift parameters, after which only one convolution layer needs to be run.

Though it seems more difficult for the network to adapt its behavior just with this single deep layer, we demonstrate its generality and effectiveness for different operators in the experiment section. Such a network design is able to outperform the state-of-the-art competitors \cite{chen2016bilateral,gharbi2017deep} by a large margin. More comprehensive ablation study about the learned layer type (convolution or instance normalization) and position $k$ is conducted in the analysis section.

\section{Experiments on the proposed framework}\label{experiment:proposed}

To demonstrate the ability of our proposed framework in incorporating parameterised image operators while maintaining their accuracy, we evaluate the proposed decoupled learning framework with different training configurations as shown from subsection \ref{experiment:single_operator} to \ref{experiment:multi_operators}. We leverage two representative types of image processing tasks: image filtering and image restoration. Within each of them, more than four popular operators are selected for detailed experiments. Below, we briefly introduce all the operators and their implementation details as follows.

\subsection{Choice of Image Operators}

{\textbf{Image Filtering:}} here we employ six popular image filters, denoted as $L_0$ \cite{xu2011image}, WLS \cite{farbman2008edge}, RTV \cite{xu2012structure}, RGF \cite{zhang2014rolling}, WMF \cite{zhang2014100} and LLF \cite{aubry2014fast}, which have been developed specifically for many different applications, such as edge-preserving image smoothing, texture removal, detail exaggeration, image abstraction, and image enhancement.
\begin{itemize}
	\item \textbf{$L_0$ smooth} \cite{xu2011image} - sharpening major image structures while eliminating a manageable degree of details by minimizing $L_0$ image gradients.
	\item \textbf{RTV} \cite{xu2012structure} - extracting structures from textures by optimizing the new inherent variation and relative total variation measures.
	\item \textbf{WLS} \cite{farbman2008edge}· - constructing edge-preserving multi-scale image decompositions for progressive coarsening of images.
	\item \textbf{RGF} \cite{zhang2014rolling} - removing textures by controlling detail smoothing under a scale measure.
	\item \textbf{WMF} \cite{zhang2014100} - an efficient 100+ times faster weighted median filter.
	\item \textbf{LLF} \cite{aubry2014fast} - fast local Laplacian filters for tone mapping.
\end{itemize}

\noindent{\textbf{Image Restoration:}} The goal of image restoration is to recover a clear image from a corrupted image. In this paper we deal with four representative tasks in this venue: super resolution \cite{dong2014learning}, denoising \cite{mao2016image}, deblocking \cite{dong2015compression} and derain \cite{fu2017removing}, which have been studied with deep learning based approaches extensively.  Except for the derain task, all the others are tested on various parameter settings that indicate the corruption level of the input image.
\begin{itemize}
	\item \textbf{super resolution} - increasing the resolution or enhancing the lost details from a low-resolution blurry image, which is controlled by a downsampling scale with bicubic interpolation.
	\item \textbf{denoising} - restore the clear image from a noisy image, which is composed of Gaussian white noise controlled by the Gaussian standard deviation.
	\item \textbf{deblocking} - recover the image details from a compressed JPEG image differentiated by the image quality factor.
    \item \textbf{derain} - removing rain streaks from a captured rainy image.
\end{itemize}

\subsection{Implementation Details}

{\textbf{Dataset:} } We take use of the 17k natural images in the PASCAL VOC dataset as the clear images to synthesize the ground truth training samples. The PASCAL VOC images are picked from Flicker, and consists of a wide range of viewing conditions. To evaluate our performance, 100 images from the dataset are randomly picked as the test data for the image filtering task. While for the restoration tasks, we take the well-known benchmark for each specific task for testing, which is specifically BSD100 (super resolution), BSD68 (denoise), LIVE1 (deblock), RAIN12 (derain). For the filtering task, we filter the natural images with the aforementioned algorithms to produce ground truth labels. As for the image restoration tasks, the clear natural image is taken as the target image while the synthesized corrupted image is used as input.

\noindent{\textbf{Parameter Sampling:}} To make our network able to handle continuous parameters, we generate training image pairs with a much broader scope of parameter values rather than a single one. We uniformly sample parameters in either the logarithm or the linear space depending on the specific application.  If the upper bound of the parameter range is tens or even hundreds of times larger than the lower bound, the parameters are sampled in the logarithm space to balance their magnitudes, otherwise they are sampled in the linear space.

\setlength{\tabcolsep}{1pt}
\renewcommand{\arraystretch}{1}
\begin{table*}[htp]
\begin{center}
\caption{Quantitative absolute difference between the network trained with a \textit{single} parameter value and \textit{numerous} random values for each image operator.}
\label{table:pami_1}

\begin{tabular}{c  cccc  cccc  cccc  cccc  cccc  cccc}
\toprule[0.08em]
 & \multicolumn{4}{ c }{{$L_0$}} & \multicolumn{4}{ c }{{WLS}} & \multicolumn{4}{ c }{{RTV}} & \multicolumn{4}{ c }{{RGF}} & \multicolumn{4}{ c }{{WMF}} & \multicolumn{4}{ c }{{LLF}} \\
\cmidrule{1-25}
 metric & $\lambda$  & single & nume. & diff. &  $\lambda$  & single & nume. & diff.  &  $\lambda$  & single & nume. & diff. & $\lambda$  & single & nume. & diff. &  $\lambda$  & single & nume. & diff.  &  $\lambda$  & single & nume. & diff.\\
\cmidrule{1-25}
\multirow{6}{*}{\small{PSNR}}
&0.002 & 40.69 & 39.46 & 1.23 & 0.100 & 44.00 & 42.12 & 1.88 & 0.002 & 41.11 & 40.66 & 0.45 &1.00 & 41.77 & 37.03 & 4.74 &1.00 & 39.06 & 36.79 & 2.27 &2 & 38.00 & 37.83 & 0.17\\
&0.004 & 38.96 & 38.72 & 0.24 & 0.215 & 43.14 & 42.64 & 0.50 & 0.004 & 40.91 & 41.10 & 0.19 &3.25 & 38.36 & 38.27 & 0.09 &3.25 & 39.78 & 38.76 & 1.02 &3 & 34.64 & 35.71 & 1.07\\
&0.020 & 36.07 & 35.71 & 0.36 & 1.000 & 41.93 & 41.63 & 0.30 & 0.010 & 40.50 & 41.07 & 0.57 &5.50 & 38.11 & 38.35 & 0.24 &5.50 & 39.94 & 38.53 & 1.41 &5 & 32.34 & 32.29 & 0.05\\
&0.093 & 33.08 & 31.92 & 1.16 & 4.641 & 39.42 & 39.64 & 0.22 & 0.022 & 41.07 & 40.77 & 0.30 &7.75 & 37.65 & 37.99 & 0.34 &7.75 & 40.06 & 39.20 & 0.86 &7 & 30.11 & 29.91 & 0.20\\
&0.200 & 31.75 & 30.43 & 1.32 & 10.00 & 39.13 & 38.51 & 0.62 & 0.050 & 40.73 & 39.18 & 1.55 &10.0 & 37.52 & 37.08 & 0.44 &10.0 & 39.49 & 38.72 & 0.77 &8 & 29.53 & 28.95 & 0.58\\
\cmidrule{2-25}
&ave.  & 36.11 & 35.25 & \textbf{0.86} & ave.  & 41.52 & 40.91 & \textbf{0.61} & ave.  & 40.86 & 40.55 & \textbf{0.31} &ave.  & 38.68 & 37.74 & \textbf{0.93} & ave.  & 39.66 & 38.40 & \textbf{1.26}  & ave. & 32.93 & 32.94 & \textbf{0.01} \\
\cmidrule{1-25}
\multirow{6}{*}{\small{SSIM}}
&0.002 & 0.989 & 0.988 & 0.001 & 0.100 & 0.994 & 0.993 & 0.001 & 0.002 & 0.987 & 0.988 & 0.001 &1.00  & 0.994 & 0.981 & 0.013 &1.00 & 0.985 & 0.972 & 0.013 &2 & 0.992 & 0.992 & 0 \\
&0.004 & 0.986 & 0.987 & 0.001 & 0.215 & 0.993 & 0.993 & 0     & 0.004 & 0.989 & 0.990 & 0.001 &3.25  & 0.986 & 0.986 & 0     &3.25 & 0.985 & 0.979 & 0.006 &3 & 0.988 & 0.990 & 0.02 \\
&0.020 & 0.982 & 0.981 & 0.001 & 1.000 & 0.992 & 0.991 & 0.001 & 0.010 & 0.990 & 0.991 & 0.001 &5.50  & 0.985 & 0.986 & 0.001 &5.50 & 0.986 & 0.981 & 0.005 &5 & 0.983 & 0.984 & 0.01\\
&0.093 & 0.977 & 0.973 & 0.004 & 4.641 & 0.987 & 0.989 & 0.002 & 0.022 & 0.992 & 0.992 & 0     &7.75  & 0.984 & 0.985 & 0.001 &7.75 & 0.986 & 0.985 & 0.001 &7 & 0.977 & 0.977 & 0\\
&0.200 & 0.973 & 0.968 & 0.005 & 10.00 & 0.986 & 0.987 & 0.001 & 0.050 & 0.992 & 0.990 & 0.002 &10.0  & 0.984 & 0.982 & 0.002 &10.0 & 0.986 & 0.984 & 0.002 &8 & 0.976 & 0.974 & 0.02\\
\cmidrule{2-25}
&ave.  & 0.981 & 0.979 & \textbf{0.002} & ave.  & 0.990 & 0.990 & \textbf{0}     & ave.  & 0.990 & 0.990 & \textbf{0} &ave.  & 0.986 & 0.984 & \textbf{0.002} & ave.  & 0.985 & 0.980 & \textbf{0.005} &ave. & 0.983 & 0.983 & \textbf{0}\\
\bottomrule
\end{tabular}
\end{center}
\end{table*}

\setlength{\tabcolsep}{1.0pt}
\renewcommand{\arraystretch}{1}
\begin{table}[t]
\begin{center}

\caption{Quantitative absolute difference in PSNR (above) and SSIM (bottom) between the network trained on a \textit{single} parameter value and \textit{numerous} random values on the three image restoration tasks. Their parameters specifically mean downsampling scale ($s$), Gaussian standard deviation ($\sigma$) and JPEG quality ($q$).}
\label{table:pami_2}

\begin{tabular}{ cccc  cccc  cccc }
\toprule[0.08em]
  \multicolumn{4}{ c }{{Super Resolution}} & \multicolumn{4}{ c }{{Denoising}} & \multicolumn{4}{ c }{{Deblock}} \\
\cmidrule{1-12}
  $s$  & single & nume. & diff. &  $\sigma$  & single & nume. & diff.  &  $q$  & single & nume. & diff. \\
\cmidrule{1-12}

 2 & 31.78 & 31.62 & 0.16 & 15 & 31.17 & 31.07 & 0.10 & 10 & 29.26 & 29.17 & 0.09 \\
 3 & 28.78 & 28.76 & 0.02 & 25 & 28.94 & 28.98 & 0.04 & 20 & 31.49 & 31.43 & 0.06 \\
 4 & 27.31 & 27.31 & 0    & 50 & 26.22 & 26.14 & 0.08 &  \\
\cmidrule{1-12}
ave.  & 29.29 & 29.23 & \textbf{0.06} & ave.  & 28.77 & 28.73 & \textbf{0.04} & ave.  & 30.37 & 30.30 & \textbf{0.07} \\
\cmidrule{1-12}

 2 & 0.894 & 0.892 & 0.002 & 15 & 0.881 & 0.883 & 0.002 & 10 & 0.817 & 0.817 & 0 \\
 3 & 0.798 & 0.796 & 0.002 & 25 & 0.821 & 0.822 & 0.001 & 20 & 0.881 & 0.882 & 0.001 \\
 4 & 0.728 & 0.726 & 0.002 & 50 & 0.722 & 0.718 & 0.004 & \\
\cmidrule{1-12}
ave.  & 0.806 & 0.804 & \textbf{0.002} & ave.  & 0.808 & 0.807 & \textbf{0.001} & ave.  & 0.849 & 0.849 & \textbf{0} \\
\bottomrule
\end{tabular}
\end{center}
\end{table}

\subsection{Results on the Single Parameterized Operator}\label{experiment:single_operator}

\textbf{Image Filtering. } We first evaluate the performance of six image operators with various controllable parameters individually. We train one network for each parameter value ($\lambda$) in one operator, and also train a network jointly on continuous random values sampled from the operator's parameter range, which can be inferred from the $\lambda$ column in Table \ref{table:pami_1}. The performance of the two networks is evaluated on the test dataset with PSNR and SSIM error metrics. Since our goal is to measure the performance difference between these two strategies, we directly compute the absolute difference of their errors and demonstrate the results in Table \ref{table:pami_1}.

As can be seen from the table, though our proposed framework trained with numerous parameter settings lags a little behind the one trained on a single parameter value, their difference is very small especially for the visually more important error metric SSIM. For each image operator, previous methods usually requires to train separate networks for each parameter value, while our proposed approach only trains one single network jointly. Moreover, these image operators are dedicated to different image processing applications, the proposed framework is still able to learn all of them well, which verifies the versatility and robustness of our strategy.

\begin{figure*}[htp]
\begin{center}

\begin{tabular}{cc ccccc}

$\lambda$& input &0.002 & 0.004 & 0.020 & 0.093 & 0.200
\\

\raisebox{2cm}{\rotatebox[origin=c]{90}{\footnotesize{{$L_0$}}}}
&\includegraphics[width=2.90cm]{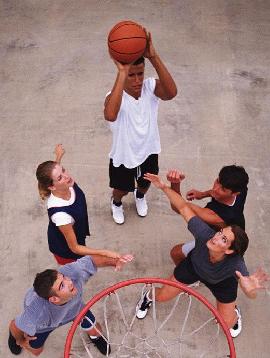}
&\includegraphics[width=2.90cm]{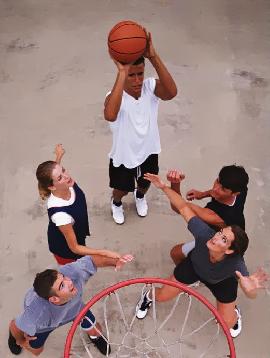}
&\includegraphics[width=2.90cm]{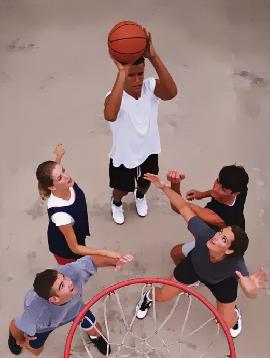}
&\includegraphics[width=2.90cm]{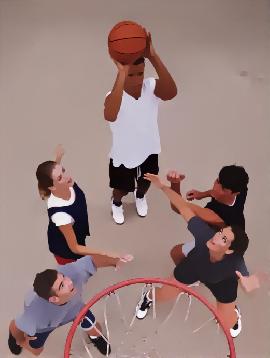}
&\includegraphics[width=2.90cm]{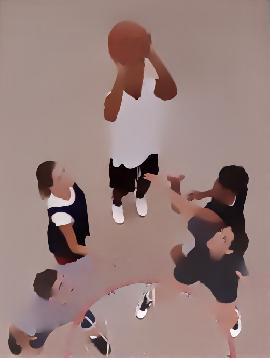}
&\includegraphics[width=2.90cm]{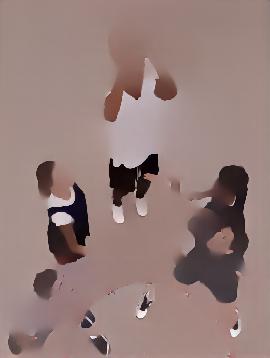}
\\

$\lambda$& input &0.002 & 0.04 & 0.010 & 0.022 & 0.050
\\

\raisebox{1.1cm}{\rotatebox[origin=c]{90}{\footnotesize{{RTV}}}}
&\includegraphics[width=2.90cm]{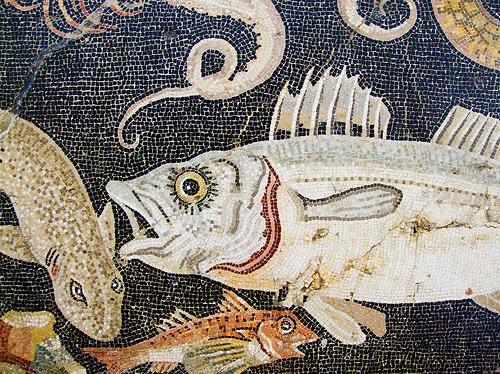}
&\includegraphics[width=2.90cm]{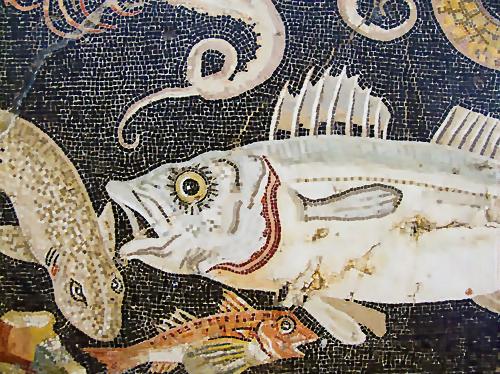}
&\includegraphics[width=2.90cm]{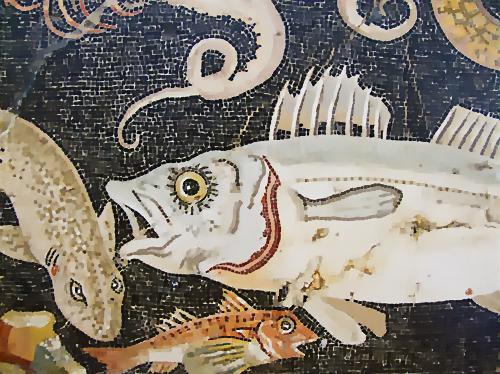}
&\includegraphics[width=2.90cm]{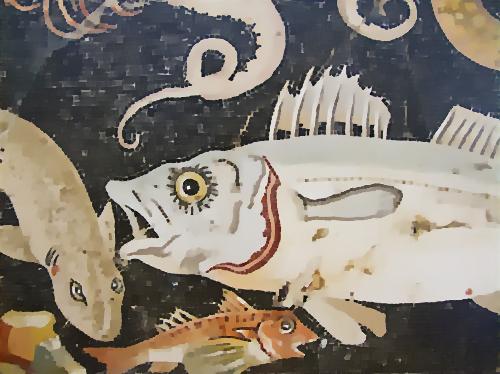}
&\includegraphics[width=2.90cm]{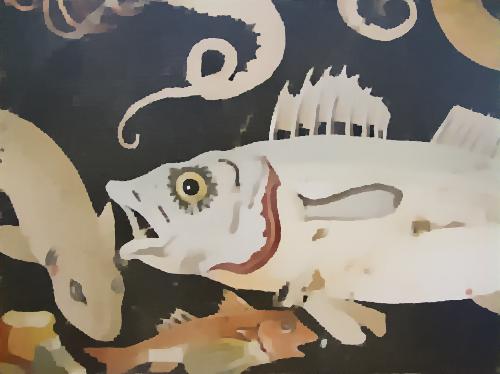}
&\includegraphics[width=2.90cm]{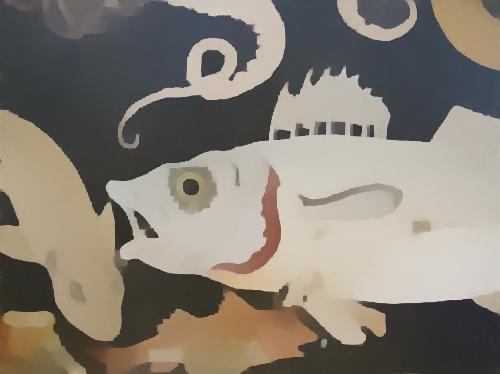}
\\

$\lambda$& input &1.00 & 3.25 & 5.50 & 7.75 & 10.00
\\

\raisebox{1.35cm}{\rotatebox[origin=c]{90}{\footnotesize{{RGF}}}}
&\includegraphics[width=2.90cm]{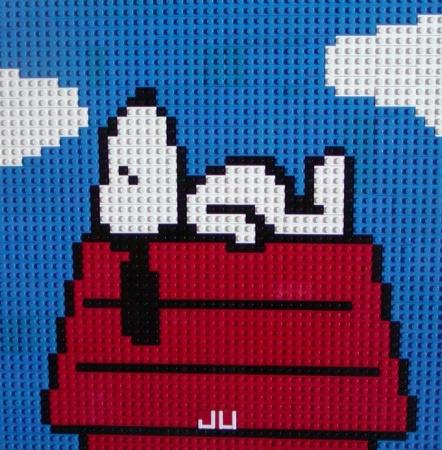}
&\includegraphics[width=2.90cm]{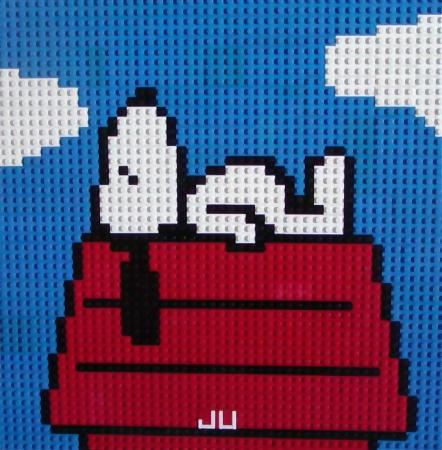}
&\includegraphics[width=2.90cm]{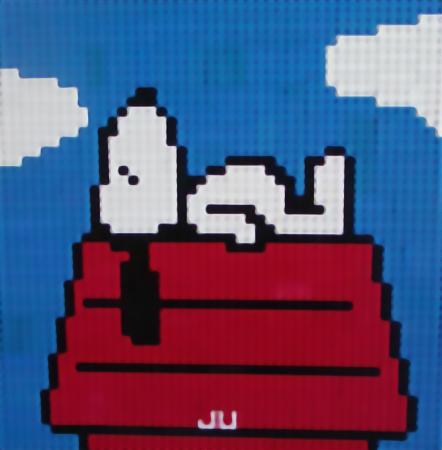}
&\includegraphics[width=2.90cm]{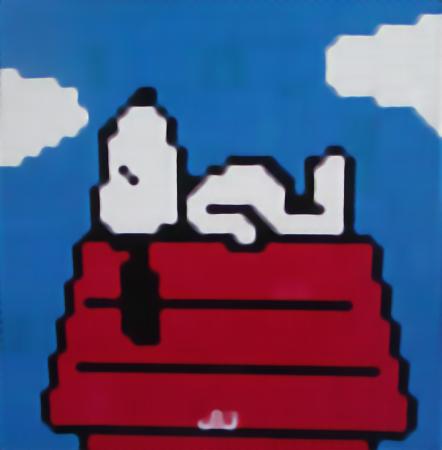}
&\includegraphics[width=2.90cm]{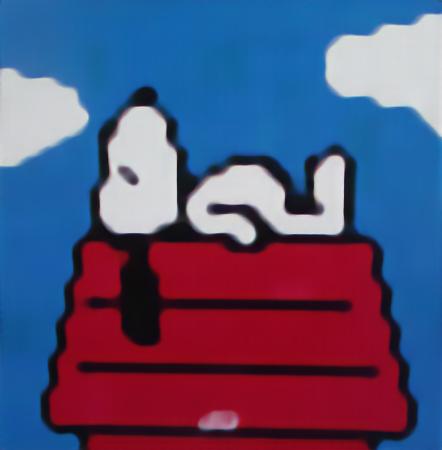}
&\includegraphics[width=2.90cm]{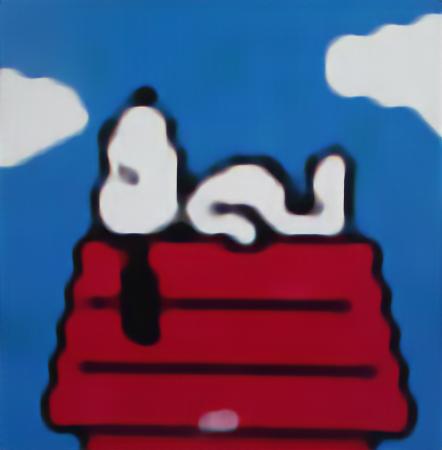}
\\

$\lambda$& input &1.00 & 3.25 & 5.50 & 7.75 & 10.00
\\

\raisebox{0.9cm}{\rotatebox[origin=c]{90}{\footnotesize{{WMF}}}}
&\includegraphics[width=2.90cm]{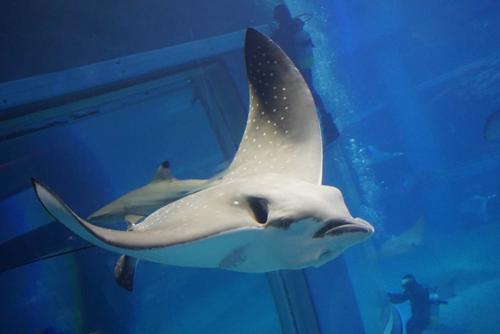}
&\includegraphics[width=2.90cm]{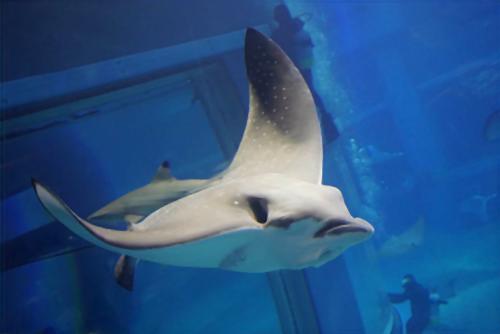}
&\includegraphics[width=2.90cm]{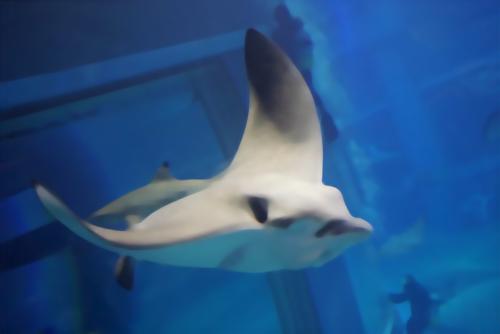}
&\includegraphics[width=2.90cm]{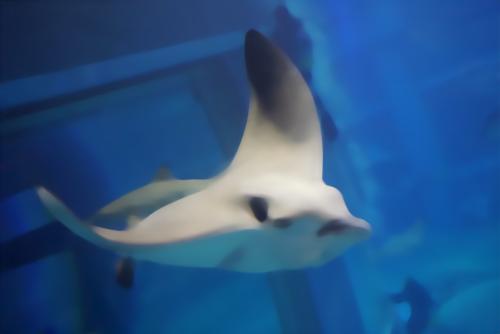}
&\includegraphics[width=2.90cm]{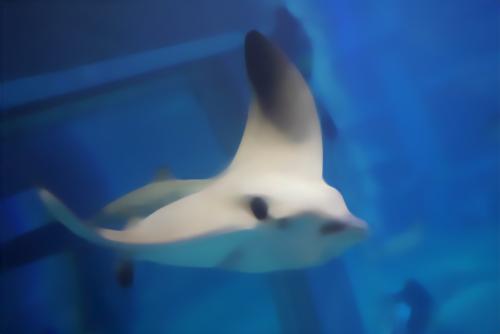}
&\includegraphics[width=2.90cm]{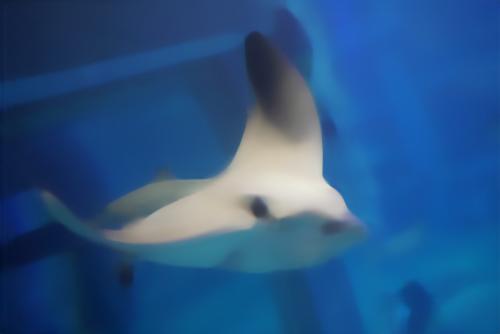}
\\

$\lambda$& input &2 & 3 & 5 & 7 & 8
\\

\raisebox{1.9cm}{\rotatebox[origin=c]{90}{\footnotesize{{LLF}}}}
&\includegraphics[width=2.90cm]{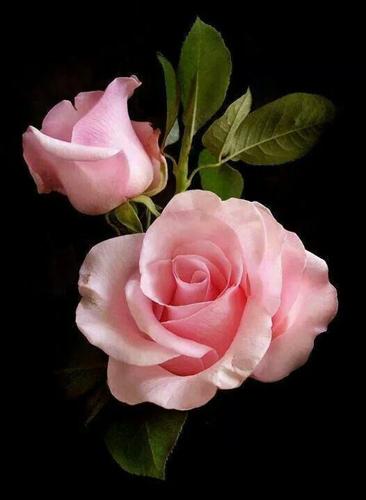}
&\includegraphics[width=2.90cm]{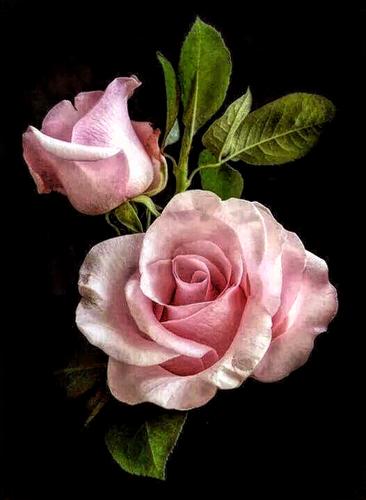}
&\includegraphics[width=2.90cm]{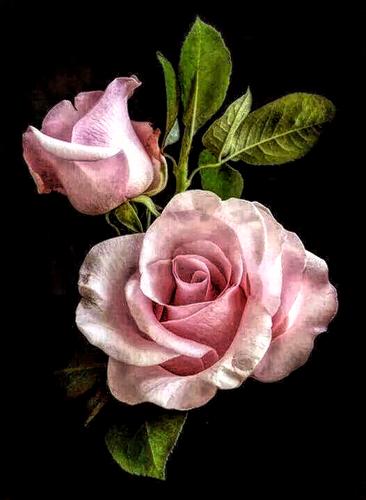}
&\includegraphics[width=2.90cm]{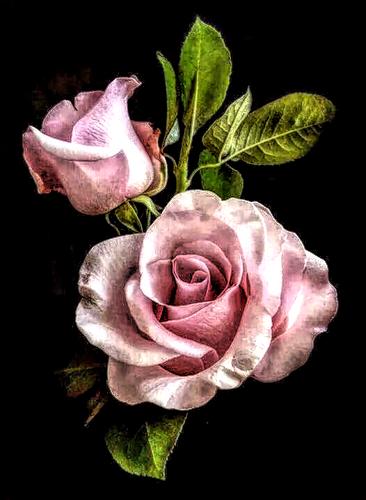}
&\includegraphics[width=2.90cm]{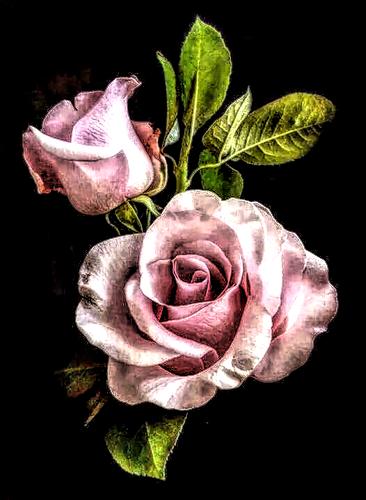}
&\includegraphics[width=2.90cm]{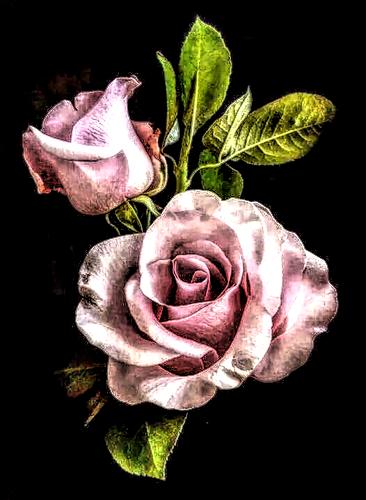}
\\

\end{tabular}

\end{center}
\vspace{-2mm}
\caption{Visual examples produced by our framework trained on continuous parameter settings of six image filters independently. Note all the visual effects for one filter are generated by a single network.}
\vspace{-2mm}
\label{figure:pami_1}
\end{figure*}

Some visual results of our proposed framework are shown in \Fref{figure:pami_1}. As can be seen, our single network trained on continuous random parameter values is capable of predicting high-quality smooth images of various strengths.

\setlength{\tabcolsep}{5pt}
\renewcommand{\arraystretch}{1}
\begin{table*}[htp]
\begin{center}

\caption{Numerical results of our proposed framework jointly trained over different number of image operators (\#operators). ``6/4'' refers to the results jointly trained over either the front 6 filtering based approaches or the last 4 restoration tasks. ``10" is the results of jointly training all 10 tasks. Our approach also achieves superior performance over one baseline \cite{chen2017fast} on all the 10 image operators.}
\label{table:pami_4}

\begin{tabular}{ cc c cccccccccc c }
\toprule[0.08em]
metric & method & $\#$ope. & $L_0$ & WLS & RTV & RGF & WMF & LLF & SR & denoise & deblock & derain & average\\
\cmidrule{1-14}
\multirow{4}{*}{\small{PSNR}} & \multirow{3}{*}{\small{Ours}} & 1 & 35.25 & 40.91 & 40.55 & 37.74 & 38.40 & 32.94 & 29.23 & 28.73 & 30.30 & 29.86 & \textbf{34.40} \\
&& 6/4  & 33.27 & 37.39 & 37.00 & 35.41 & 36.06 & 30.08 & 28.89 & 28.67 & 30.10 & 30.32 & \textbf{32.72} \\
&& 10   & 32.67 & 36.59 & 36.03 & 34.64 & 35.08 & 29.77 & 28.53 & 28.36 & 29.69 & 30.45 & \textbf{32.18} \\
\cmidrule{2-14}
& \cite{chen2017fast} & 10   & 31.01 & 34.85 & 34.20 & 33.10 & 33.61 & 28.58 & 28.21 & 28.05 & 29.48 & 29.12 & \textbf{31.02} \\
\cmidrule{1-14}
\multirow{4}{*}{\small{SSIM}} & \multirow{3}{*}{\small{Ours}} & 1    & 0.979 & 0.991 & 0.990 & 0.984 & 0.980 & 0.984 & 0.804 & 0.807 & 0.849 & 0.893 & \textbf{0.926} \\
&& 6/4  & 0.969 & 0.980 & 0.979 & 0.974 & 0.967 & 0.976 & 0.797 & 0.800 & 0.842 & 0.893 & \textbf{0.918} \\
&& 10   & 0.965 & 0.978 & 0.975 & 0.969 & 0.962 & 0.971 & 0.789 & 0.789 & 0.837 & 0.895 & \textbf{0.913} \\
\cmidrule{2-14}
& \cite{chen2017fast} & 10   & 0.949 & 0.969 & 0.964 & 0.953 & 0.947 & 0.961 & 0.779 & 0.777 & 0.834 & 0.863 & \textbf{0.899} \\
\bottomrule
\end{tabular}
\end{center}
\vspace{-10mm}
\end{table*} 
\setlength{\tabcolsep}{1pt}

\noindent\textbf{Image Restoration. } We then evaluate the proposed framework on three popular image restoration tasks as shown in Table \ref{table:pami_2}, which perform essentially different from image filtering. Our model learns to recover from different corrupted input images, instead of learning to obtain different visual effects given the same input image as in the image filtering task.

\begin{figure*}[t]
\begin{center}

\begin{tabular}{c ccccccc}

$s$ & ground truth & input & 2 & input & 3 & input & 4
\\

\raisebox{0.8cm}{\rotatebox[origin=c]{90}{\footnotesize{{SR}}}}
&\includegraphics[width=2.5cm]{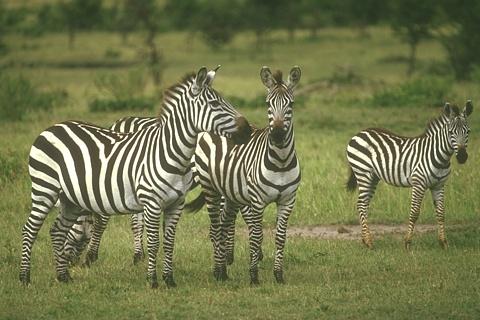}
&\includegraphics[width=2.5cm]{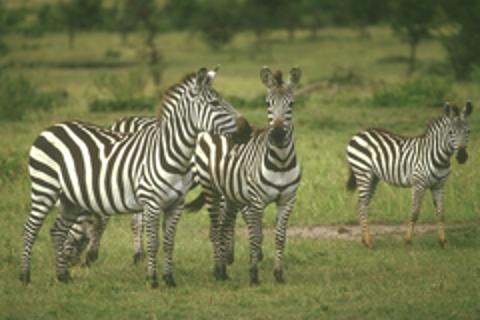}
&\includegraphics[width=2.5cm]{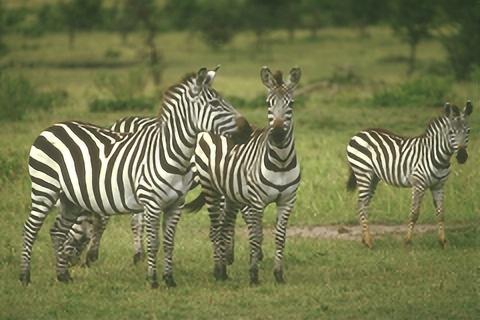}
&\includegraphics[width=2.5cm]{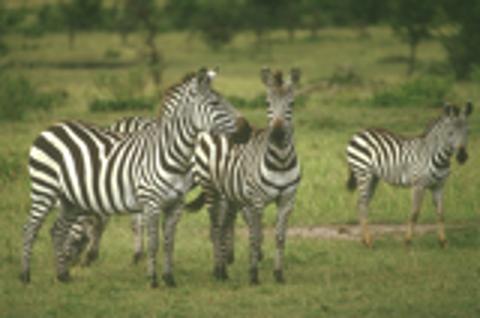}
&\includegraphics[width=2.5cm]{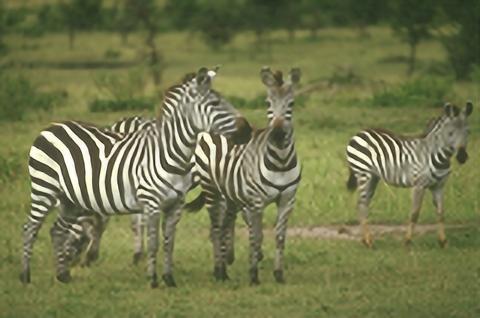}
&\includegraphics[width=2.5cm]{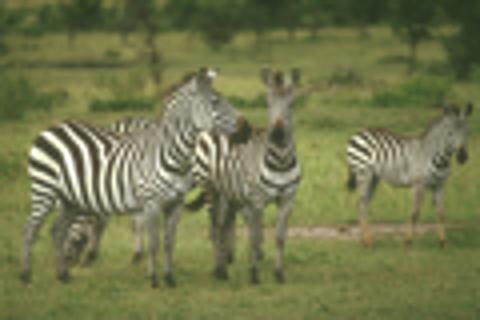}
&\includegraphics[width=2.5cm]{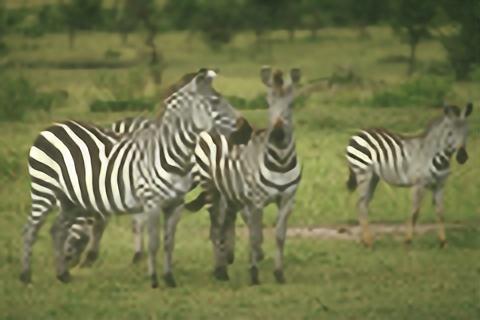}
\\

$\sigma$ & ground truth & input & 15 & input & 25 & input & 50
\\

\raisebox{1.7cm}{\rotatebox[origin=c]{90}{\footnotesize{{denoise}}}}
&\includegraphics[width=2.5cm]{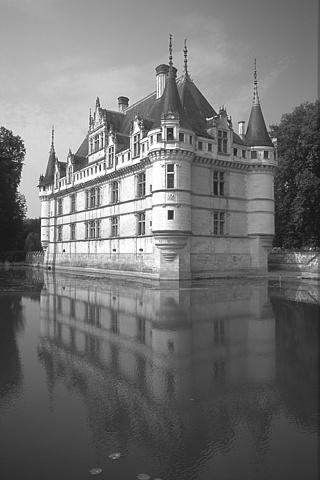}
&\includegraphics[width=2.5cm]{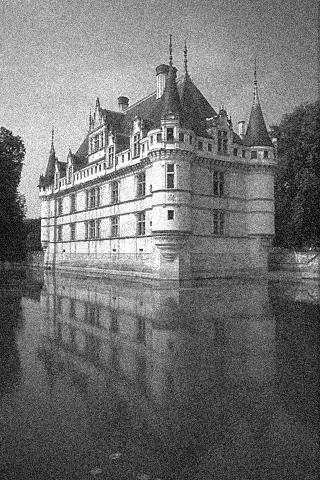}
&\includegraphics[width=2.5cm]{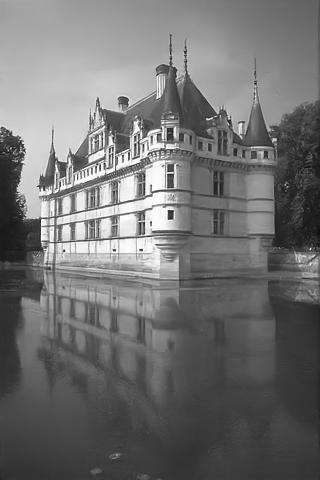}
&\includegraphics[width=2.5cm]{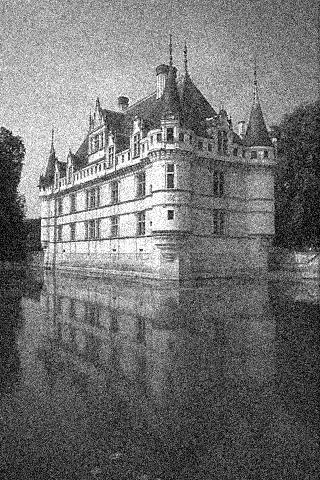}
&\includegraphics[width=2.5cm]{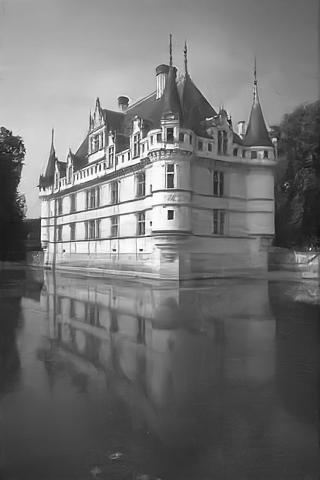}
&\includegraphics[width=2.5cm]{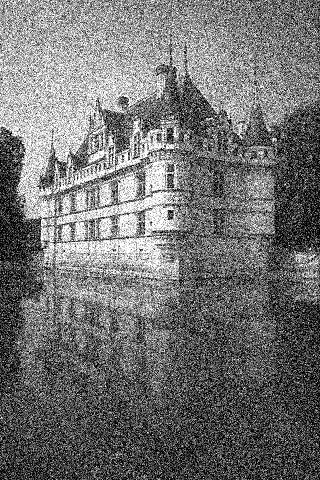}
&\includegraphics[width=2.5cm]{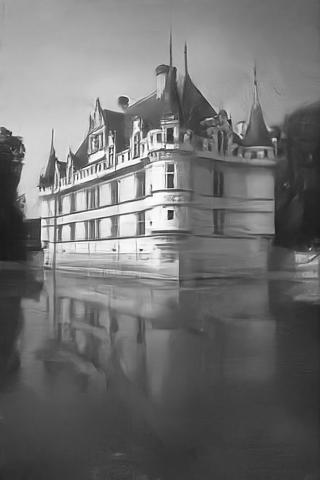}
\\

$q$ & ground truth & input & 10 & input & 20 &  &
\\

\raisebox{1.8cm}{\rotatebox[origin=c]{90}{\footnotesize{{deblock}}}}
&\includegraphics[width=2.5cm]{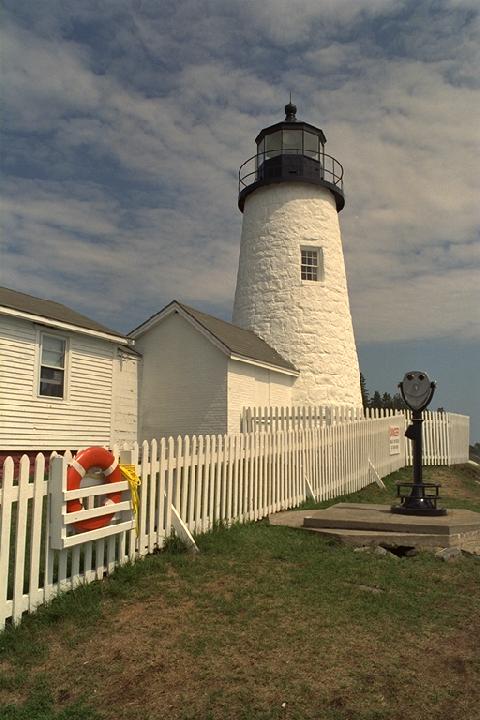}
&\includegraphics[width=2.5cm]{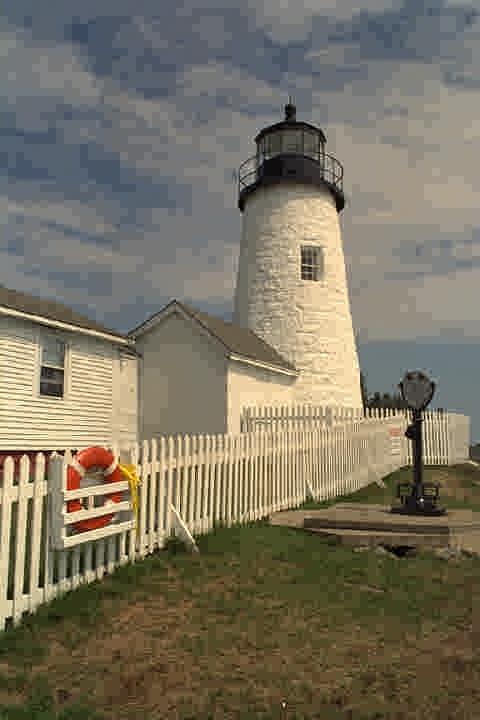}
&\includegraphics[width=2.5cm]{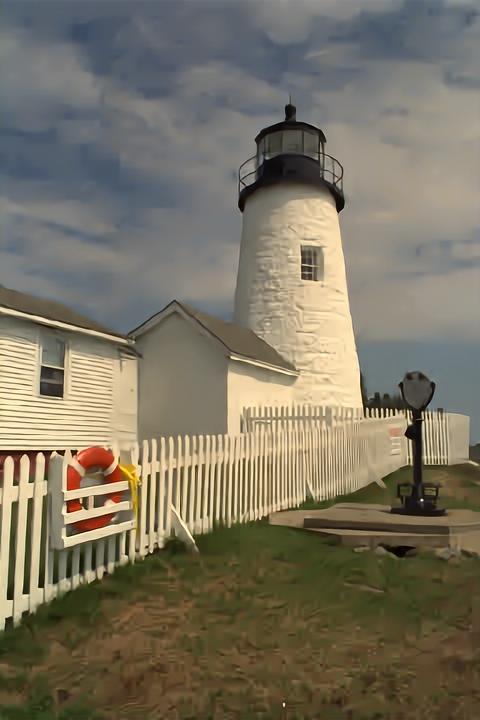}
&\includegraphics[width=2.5cm]{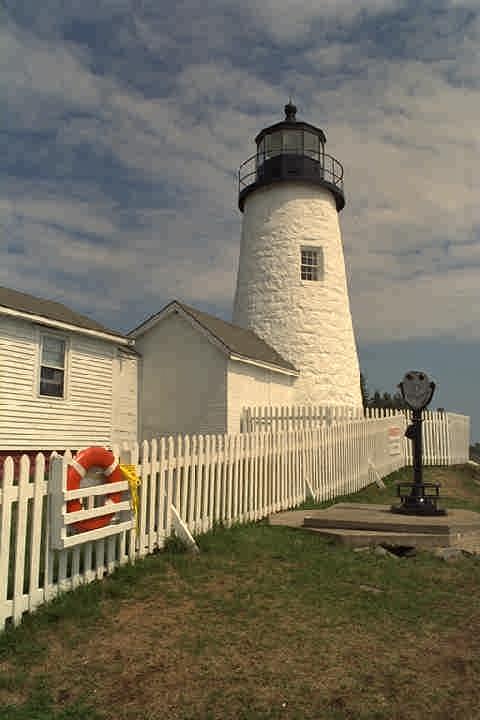}
&\includegraphics[width=2.5cm]{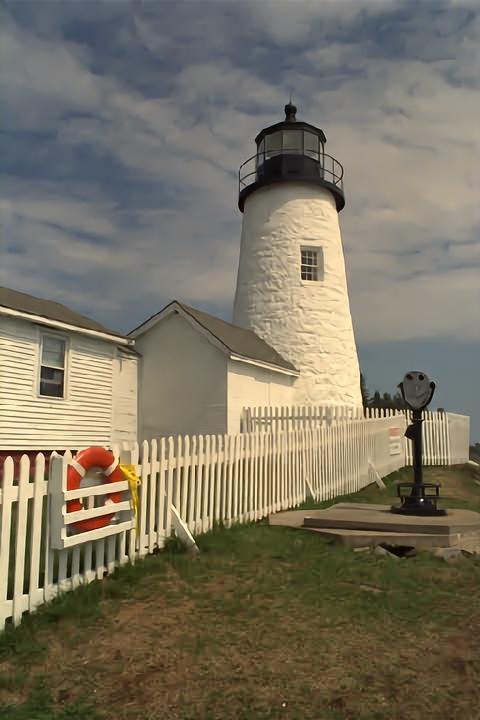}
&
&
\\
\end{tabular}

\end{center}
\vspace{-2mm}
\caption{Visual examples produced by our framework trained on continuous parameter settings of three image restoration tasks independently. Note all the clean images for one restoration task are generated by a single network.}
\label{figure:pami_2}
\vspace{-4mm}
\end{figure*}

As shown in Table \ref{table:pami_2}, our results trained jointly on continuous random parameter values also show no big difference from the one trained solely on an individual parameter value, which further validate our algorithm in a broader image processing literature. Some corresponding visual results of our proposed framework are shown in Figure \ref{figure:pami_2}.
\setlength{\tabcolsep}{1pt}

\subsection{Results on Jointly Training Multiple Image Operators} \label{experiment:multi_operators}

Intuitively, another challenging case for our proposed framework is to incorporate multiple distinct image operators into a single learned neural network, which is much harder to be trained due to their different implementation details and purposes. To explore the potential of our proposed neural network, we experiment by jointly training over (\textit{i}). 6 filtering based operators, (\textit{ii}). 4 image restoration operators or (\textit{iii}). all the 10 different operators altogether. To generate training images of each image operator, we sample random parameter values continuously within its parameter range. Since there is no parameter tuning for the derain task, we leverage its default parameter setting for training.

The input to the \textit{weight learning} network now takes two parameters, one indicates the specific image operator while the other is the random parameter values assigned to the specified filter. These 10 image operators are denoted simply by 10 discrete values that range from 0.1 to 1.0 in the input parameter vector. Since the absolute parameter range may differ a lot from operator to operator, for example, [2,4] for super resolution and [0.002,0.2] for $L_0$ filter, we rescale the parameters in all the operators into the same numerical range to enable consistent back-propagated gradient magnitude.

As shown in Table \ref{table:pami_4}, training on each individual image operator achieves the highest numerical score (\#ope.=1), which is averaged over multiple different parameter settings just like in previous tables. While jointly training over either 6 image filters or 4 restoration tasks (\#ope.=6/4), even for the case where all 10 image operators are jointly trained (\#ope.=10), their average performance degrades but still achieves close results to the best score. It means with the same network structure, our framework is able to incorporate all these different image operators together into a single network without losing much accuracy.


\subsection{Comparison with Baseline \cite{chen2017fast}}
We compare our proposed framework with one naive approach that employs only the \textit{base} network with additional input channels as \cite{chen2017fast}, which indicates the parameter values and image operators separately. Each additional channel is occupied with a single value.

The results are shown in Table \ref{table:pami_4}, which trains ten different image operators including both image filtering and restoration tasks together. We can see that the baseline from \cite{chen2017fast} lags behind us across all ten image operators. The potential reason for this phenomenon could be that it is more difficult to learn unified convolution weights to be suitable for tasks different in both goals and implementations. By contrast, the convolutional weights of our \textit{base} network are adaptively changed for different tasks and input parameters. Theoretically speaking, our learned network should have the capability to express the base network with more numerous possibilities.


\setlength{\tabcolsep}{3pt}
\renewcommand{\arraystretch}{1}
\begin{table}[t]
	\begin{center}
		\caption{Quantitative evaluation (PSNR) on the higher dimensional parameter space of the RTV filter. ``$\#$dim.'' indicates the number of leveraged parameter dimensions, and ``$\#$ope.'' is the number of jointly trained image operators.}
		\label{table:pami_3}
		\begin{tabular}{ cc cccccc }
			\toprule[0.08em]
			&& \multicolumn{6}{ c }{{$\lambda$ (RTV)}} \\
			\cmidrule{3-8}
			$\#$ope. & $\#$dim. & 0.002 & 0.004 & 0.010 & 0.022 & 0.050 & average \\
			\cmidrule{1-8}
			1 & 1 & 40.66 & 41.10 & 41.07 & 40.77 & 39.18 & \textbf{40.55} \\
			1 & 2 & 40.70 & 40.89 & 40.78 & 40.56 & 39.23 & \textbf{40.43} \\
			1 & 3 & 40.19 & 40.82 & 40.95 & 40.93 & 39.78 & \textbf{40.53} \\
			1 & 4 & 40.18 & 40.83 & 40.95 & 40.87 & 39.68 & \textbf{40.50} \\
			\cmidrule{1-8}
			10 & 2 & 37.44 & 37.16 & 36.50 & 35.51 & 33.52 & \textbf{36.03} \\
			\bottomrule
		\end{tabular}
	\end{center}
\vspace{-5mm}
\end{table}

\subsection{Generalization to Higher Dimensional Parameter Space}\label{experiment:multi_parameter}

Except for experimenting on a single input parameter, we also demonstrate the experiments on inputting multiple types of parameters that belongs to either one image operator or various operators.

In this section, we evaluate the performance on the famous texture removal tool RTV \cite{xu2012structure}. RTV has multiple parameters to adjust its visual effects, such as $\lambda$ which balances between the data prior term and smoothness term in its energy function, $\sigma$ which controls the spatial scale for computing the windowed variation and is even more effective in removing textures, $\epsilon_s$ that controls the sharpness of the final results and $maxIter$ that refers to the number of iterations in the optimization.

We train our network on the continuous parameter range of $\lambda$ first, therefore the input parameter is a one-element vector and constrained in the one dimensional space. Its results are shown in Table \ref{table:pami_3} with $\#$ope. and $\#$dim. both equal to 1.  Then we add one more parameter at a time to increase the input parameter dimension until all the aforementioned four parameters are included to form a four-element input vector, where $\#$dim. equals to 4. Moreover, following the experimental setting in Table \ref{table:pami_4}, we also train our network on 10 different image operators including the balance weight $\lambda$ for RTV to form a two dimensional parameter space. We evaluate the performance on 5 discrete parameter values of $\lambda$ that is commonly adopted for all the parameter dimensions. 

We can see that when focusing on a single image operator (RTV) with up to even 4 parameters, the resultant performance of different parameter values are similar to each other in a reasonable range. But when we incorporate more image operators into joint training, the performance on RTV filter decreases significantly.  A reasonable explanation for this phenomenon is that as the visual effects introduced by the additional input parameter are more different from the existing one, it becomes harder to incorporate the additional input parameter for joint training.  Regarding the experiment in Table \ref{table:pami_3}, the additional parameter for more image operators introduces more different visual effects compared to the four inherent parameters in the RTV filter, and hence becomes harder for joint training.

Note while incorporating more parameter dimensions into the training process, we don't modify the amount of training samples and iterations. Each training sample is generated by randomly sampling the leveraged parameters.

\section{Experiments on the cheap parameter-tuning version}\label{sec:real-time}

In this section, we evaluate the cheap parameter-tuning adaption of our proposed framework with many recent filtering and restoration approaches. Most compared approaches design some complex and advanced deep neural network to achieve the optimal performance for their specific dedicated tasks. In order to balance between the performance and computation costs, we slightly modify the structure of $\mathcal{N}_{base}$ by utilizing the depth-wise convolution \cite{chollet2017xception} and increasing the intermediate feature channels (64 to 128) in this section. We validate and study the effectiveness of the proposed framework, which not only runs efficiently for parameter tuning but also achieves state-of-the-art performance compared to many recent deep learning approaches.

\subsection{Image Filtering}
We compare our proposed framework with two recent state-of-the-art approaches \cite{chen2016bilateral,gharbi2017deep}, which share the closest spirit as ours in reproducing the image filters as well as possible while running in real-time by varying between different image operators.

To conduct on a broader scope of image operators like the other two approaches, besides the aforementioned six image filters, we further add four operators with more different visual effects for a fair comparison as follows.
\begin{itemize}
\item \textbf{LLF remapping} \cite{aubry2014fast} - fast local Laplacian filters for tone mapping by leveraging a remapping function.
\item \textbf{WLS enhancement} \cite{farbman2008edge} - detail exaggeration implemented via multi-scale image decompositions.
\item \textbf{Stylization} \cite{aubry2014fast} - transfer the “look” of one photographer’s masterpiece onto another photo via fast local Laplacian filters.
\item \textbf{Abstraction} \cite{lu2012combining} - pencil drawing by combining the tone and stroke structures, which complement each other in generating visually constrained results.
\end{itemize}

Note the six image filters mentioned in Section \ref{experiment:proposed} contains controllable parameters that indicate different visual effects, while each of the above four operators represents a unique fixed visual effect. All these ten image operators are fed through the networks for joint training. The input parameters are sampled similarly as in section \ref{experiment:multi_operators}.

The image operators deployed in \cite{chen2016bilateral,gharbi2017deep} are relatively diverse, which include both photographic effects \cite{aubry2014fast} and image structure manipulations \cite{l0smoothing2011} that are also demonstrated in our paper.	Even if our deployed image operators are not exactly the same as theirs \cite{chen2016bilateral,gharbi2017deep}, we believe the idea proposed in their paper is general enough, and we apply them to the same operators in our paper (some included and some not included in their papers).


In Table \ref{table:pami_5}, we demonstrate the numerical results of our proposed framework compared with the competitors BGU \cite{chen2016bilateral} and DBL \cite{gharbi2017deep}. For a fair comparison, the evaluation is conducted on one parameter setting (the default one) for each image operator following the other two approaches. As can be seen from the numerical errors (PSNR and SSIM), our results achieve significantly better results. The average PSNR over 10 image operators are about 5dB larger than the second best competitor.

Note both BGU and DBL learn to reproduce one specific operator with the default parameter setting, while \textit{our approach learns all the 10 image operators with their full range of parameter values jointly within one single network}, which is much more difficult and challenging.

\subsection{Image Restoration}
Furthermore we compare our framework on the four aforementioned image restoration tasks: super resolution, denoising, deblocking and derain with many recent restoration methods as shown from Table \ref{table:pami_6} to Table \ref{table:pami_9}. Except for the derain task, all the others are tested on various parameter settings that indicate the corruption level of the input image. Though these existing algorithms are dedicated on each specific field with specifically designed algorithms or network structures, our proposed framework is still able to achieve very competitive results on all these tasks. Need to note that all these tasks are jointly trained with one single network by our approach.

Given each new example for the restoration tasks, it needs to rerun the whole network from the beginning. Therefore it's not a very valid case to justify our cheap parameter-tuning framework. However, theoretically this experiment still demonstrates the extraordinary ability of deep networks to incorporate different restoration capabilities within a very limited parameterized operation (the last instance normalization layer).

\setlength{\tabcolsep}{3pt}
\renewcommand{\arraystretch}{1}
\begin{table}[t]
	\begin{center}
		\caption{Running time evaluation (milliseconds) of our decouple learning framework and its cheap parameter tuning module on different image resolutions, along with the baseline \cite{chen2017fast}. It's evaluated on GPU devices by default without specifications.}
		\label{table:pami_time}
		\begin{tabular}{ l cccc }
			\toprule[0.08em]
			Resolution & \cite{chen2017fast} & Ours & ParaTuning &  ParaTuning (CPU)\\
			\cmidrule{1-5}			
			VGA (640$\times$480) & 4.87 & 6.20 & 0.59 & 0.65 \\
			720p (1280$\times$720) & 5.02 & 6.45 & 0.72 & 0.85 \\
			1080p (1920$\times$1080) & 5.88 & 6.90 & 0.79 & 0.98\\
			\bottomrule
		\end{tabular}
	\end{center}
\vspace{-5mm}
\end{table}

\subsection{Running Time Comparison}

In this section, we evaluate the running time of our proposed framework and the cheap parameter tuning module on different popular image resolutions. As shown in Table \ref{table:pami_time}, given a new input, we need to run through the whole model that takes less than 10 milliseconds for a 1080p image, but while switching between different image operations it takes only less than 1 millisecond on either GPU or CPU device, which is almost ten times faster than running the full model. Our implementation leverages the multi-core CPU package (NNPACK) in the MXNet framework for acceleration, and our model runs on a 20-core CPU device.

We also evaluate the running time of the baseline \cite{chen2017fast}. Note their approach and ours share most of the \textit{base} network structure, while their parameter tuning is implemented by adding additional input channels to the \textit{base} network, which doesn't increase the computation cost much. On the other hand, our framework includes one more \textit{weight learning} network and hence lags behind them on the running time. From this table, we can also see that the \textit{weight learning} network takes around 1.02 milliseconds to process a 1080p image, which is very computationally efficient.

\setlength{\tabcolsep}{5pt}
\renewcommand{\arraystretch}{1}
\begin{table*}[htp]
\begin{center}
\caption{Quantitative comparison with state-of-the-art approaches in reproducing image operators.}
\vspace{-1mm}
\label{table:pami_5}
\begin{tabular}{cc cccccccccc c}
\toprule[0.08em]
\cmidrule{1-13}
metric & method & $L_0$ & WLS & RTV & RGF & WMF & LLF & LLF remap & WLS enhance & Stylization & Abstraction & Average \\
\cmidrule{1-13}
\multirow{3}{*}{\small{PSNR}}
&BGU & 31.76 & 27.03 & 26.15 & 22.71 & 21.27 & 26.97 & 33.05 & 26.93 & 14.31 & 16.11 & \textbf{24.62} \\
&DBL & 28.67 & 30.63 & 28.52 & 27.11 & 26.88 & 25.13 & 29.34 & 28.29 & 25.08 & 19.61 & \textbf{26.92} \\
&Ours & 31.81 & 36.59 & 34.28 & 33.29 & 34.00 & 29.74 & 32.41 & 34.40 & 26.66 & 25.61 & \textbf{31.87} \\
\cmidrule{1-13}
\multirow{3}{*}{\small{SSIM}}
&BGU & 0.912 & 0.915 & 0.848 & 0.776 & 0.765 & 0.936 & 0.978 & 0.931 & 0.673 & 0.427 & \textbf{0.816} \\
&DBL & 0.852 & 0.890 & 0.826 & 0.805 & 0.786 & 0.899 & 0.945 & 0.944 & 0.887 & 0.502 & \textbf{0.833} \\
&Ours & 0.946 & 0.971 & 0.948 & 0.945 & 0.940 & 0.967 & 0.969 & 0.986 & 0.927 & 0.835 & \textbf{0.943} \\
\bottomrule
\end{tabular}
\end{center}
\end{table*}

\setlength{\tabcolsep}{5pt}
\renewcommand{\arraystretch}{1}
\begin{table*}[htp]
\begin{center}
\caption{Quantitative results (PSNR/SSIM) of the JPEG deblocking task on the LIVE1 benchmark.}
\vspace{-1mm}
\label{table:pami_6}
\begin{tabular}{cc ccccc}
\toprule[0.08em]
Quality & JPEG & ARCNN \cite{dong2015compression} & TNRD \cite{chen2017trainable} & DnCNN \cite{zhang2017beyond} & MemNet \cite{tai2017memnet} & Ours \\
\cmidrule{1-7}
10 & 27.77/0.7730 & 28.96/0.8076 & 29.15/0.8111 & 29.19/0.8123 & 29.45/0.8193 & 29.31/0.8170 \\
20 & 30.07/0.8512 & 31.29/0.8733 & 31.46/0.8769 & 31.59/0.8802 & 31.83/0.8846 & 31.60/0.8816 \\
\bottomrule
\end{tabular}
\end{center}
\end{table*}

\setlength{\tabcolsep}{5pt}
\renewcommand{\arraystretch}{1}
\begin{table*}[htp]
\begin{center}
\caption{Quantitative results (PSNR/SSIM) of the image super resolution task on the BSD100 benchmark.}
\vspace{-1mm}
\label{table:pami_7}
\begin{tabular}{cc cccccc}
\toprule[0.08em]
Scale & Bicubic & SRCNN \cite{dong2016image} & VDSR \cite{Kim2016VDSR} & DRCN \cite{kim2016deeply} & DnCNN \cite{zhang2017beyond} & MemNet \cite{tai2017memnet} & Ours \\
\cmidrule{1-8}
2 & 29.56/0.8431 & 31.36/0.8879 & 31.90/0.8960 & 31.85/0.8942 & 31.90/0.8961 & 32.08/0.8978 & 31.67/0.8934\\
3 & 27.21/0.7385 & 28.41/0.7863 & 28.82/0.7976 & 28.80/0.7963 & 28.85/0.7981 & 28.96/0.8001 & 28.76/0.7969\\
4 & 25.96/0.6675 & 26.90/0.7101 & 27.29/0.7251 & 27.23/0.7233 & 27.29/0.7253 & 27.40/0.7281 & 27.29/0.7257\\
\bottomrule
\end{tabular}
\end{center}
\end{table*}

\setlength{\tabcolsep}{5pt}
\renewcommand{\arraystretch}{1}
\begin{table}[htp]
\begin{center}
\caption{Quantitative results (PSNR) of the image denoising task on the BSD68 benchmark.}
\vspace{-1mm}
\label{table:pami_8}
\begin{tabular}{cc cccc}
\toprule[0.08em]
$\sigma$ & BM3D \cite{dabov2007image} & WNNM \cite{gu2014weighted} & TNRD \cite{chen2017trainable} & DCDP \cite{zhang2017learning} & Ours \\
\cmidrule{1-6}
15 & 31.07 & 31.37 & 31.42 & 31.63 & 31.48 \\
25 & 28.57 & 28.83 & 28.92 & 29.15 & 29.11 \\
50 & 25.62 & 25.87 & 25.97 & 26.19 & 26.22 \\
\bottomrule
\end{tabular}
\end{center}
\end{table}

\setlength{\tabcolsep}{5pt}
\renewcommand{\arraystretch}{1}
\begin{table}[htp]
\begin{center}
\caption{Quantitative results (PSNR) of the derain task on the RAIN12 benchmark.}
\vspace{-1mm}
\label{table:pami_9}
\begin{tabular}{ccc}
\toprule[0.08em]
DerainNet \cite{fu2017clearing} & DnCNN \cite{zhang2017beyond} & Ours \\
\cmidrule{1-3}
28.94 & 30.90 & 30.08 \\
\bottomrule
\end{tabular}
\end{center}
\end{table}

\section{Understanding and analysis}

To better understand the \textit{base} network $\mathcal{N}_{base}$ and the \textit{weight learning} network $\mathcal{N}_{weight}$, we analyze some meaningful behavior behind our framework in this section.

\begin{figure*}[t]
	\includegraphics[width=0.98\linewidth]{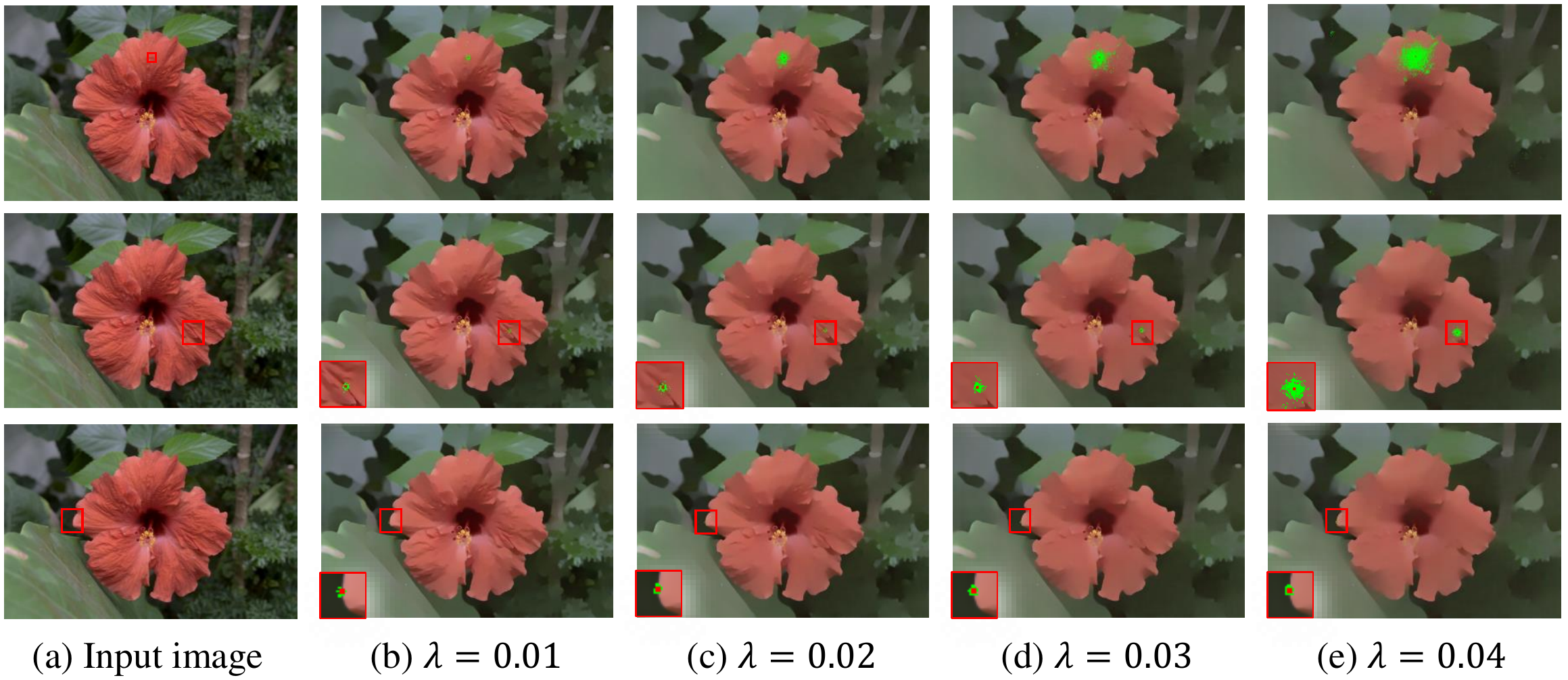}
	\caption{Effective receptive field of $L_0$ smoothing for different spatial positions and parameter $\lambda$. The top to bottom indicate the effective receptive field of a non-edge point, a moderate edge point, and a strong edge point. }
	\label{fg:receptive_field_study}
\end{figure*}

\subsection{Complexity of the Model Size} 

The saved weights of our proposed decouple learning framework come from two parts, the \textit{weight learning} network and \textit{base} network. Let's discuss our basic implementation where all the convolution weights are adaptively learned by the \textit{weight learning} network with $n$ input parameters. The \textit{base} network contains 2,432 instance normalization weights and 696,256 convolution weights to be learned. While the convolution weights are adaptively predicted by the \textit{weight learning} network, and hence are not required to be saved. For each convolution layer in the \textit{base} network, the \textit{weight learning} network contains a single fully connected layer to predict its weights. Hence the learned weights for the fc layer are totally 2,088,768 (696,256$\times$3), where $n$ equals to 2 for the case of jointly training multiple image operators. Then the total weights required to save our decouple learning framework are 2,091,200 (2,088,768+2,432).

As analyzed above, most saved weights come from the \textit{weight learning} network, which is equivalent to $n$+1 times the convolution weights in the \textit{base} network. Regardless of the detailed \textit{base} network design, such a conclusion still holds. While jointly training multiple image operators, the total saved weights of our framework are almost 3 times the weights of the independent \textit{base} network, however it implements theoretically numerous image operations defined by the continuous input parameter range. In our PyTorch implementation, the saved model for jointly training 10 image operators with two input parameters only occupies 5.33 MB, which is sufficient for embedded systems.

While adapted for cheap parameter tuning, only the weights in a single instance normalization layer are learned by the \textit{weight learning} network. The total saved weights of our framework are even closer to the one of an independent \textit{base} network, therefore it becomes more efficient to jointly train multiple image operators. As the \textit{base} network is bigger in this case, it takes 7.99 MB to save the whole model.

\subsection{The Effective Receptive Field} 

In neuroscience, the receptive field is the particular region of the sensory space in which a stimulus will modify the firing of one specific neuron. The large receptive field is also known to be  important for modern convolutional networks. Different strategies are proposed to increase the receptive field, such as deeper network structure or dilated convolution. Though the theoretical receptive field of one network may be very large, the real effective receptive field may vary with different learning targets. So how is the effective receptive field of $\mathcal{N}_{base}$ changed with different parameters $\overrightarrow{\gamma}$ and $\mathcal{I}$ ? Here we use $L_0$ smoothing \cite{l0smoothing2011} as the default example operator.

In \fref{fg:receptive_field_study}, we study the effective receptive field of a non-edge point, a moderate edge point, and a strong edge point with different smoothing parameters $\lambda$ respectively. To obtain the effective receptive field for a specific spatial point $p$, we first feed the input image into the network to get the smoothing result, then propagate the gradients back to the input while masking out the gradient of all points except $p$. Only the points whose gradient value is large than $0.025*grad_{max}$ ($grad_{max}$ is the maximum gradient value of input gradient) are considered within the receptive field and marked as green in \fref{fg:receptive_field_study}.
From \fref{fg:receptive_field_study}, we observe three important phenomena: 1) For a non-edge point, the larger the smoothing parameter $\lambda$ is, the larger the effective field is, and most effective points fall within the object boundary. 2) For a moderate edge point, its receptive field stays small until a relatively large smoothing parameter is used. 3) For a strong edge point, the effective receptive field is always small for all the different smoothing parameters. It means, on one hand, the \textit{weight learning} network $\mathcal{N}_{weight}$ can dynamically change the receptive field of $\mathcal{N}_{base}$ based on different smoothing parameters. On the other hand, the \textit{base} network $\mathcal{N}_{base}$ itself can also adaptively change its receptive field for different spatial points.

%
%

\setlength{\tabcolsep}{4pt}
\begin{table*}[t]
\begin{center}
\caption{Comparison between the statistics of convolution kernels learned with random parameter values and a single parameter value. The numbers are generated based on the WLS filter while $\lambda$ equals to 10.}
\label{table:pami_10}

\begin{tabular}{l ccccc ccccc }
\toprule[0.08em]
layer index & 1 & 2 & 3 & 4 & 5 & 6 & 7 & 8 & 9 & 10\\
\cmidrule{1-11}
correlation & 0.005 & -0.001 & 0.008 & 0.004 & -0.005 & 0.007 & 0.008 & 0.010 & -0.002 & -0.012\\
\cmidrule{1-11}
mean (nume.) & 0.219 & 0.066 & 0.004 & -0.069 & -0.227 & -0.183 & -0.214 & -0.131 & -0.152 & -0.141 \\
mean (single) & 0.542 & 0.801 & 0.604 & -0.571 & -2.090 & -0.685 & -0.520 & -1.638 & -1.604 & -1.326 \\
\cmidrule{1-11}
variance (nume.) & 15.514 & 12.166 & 15.511 & 18.231 & 17.676 & 20.408 & 16.578 & 19.420 & 16.447 & 18.292 \\
variance (single) & 542.28 & 373.70 & 490.53 & 419.55 & 482.71 & 559.00 & 532.14 & 505.67 & 471.50 & 437.69 \\
\toprule[0.08em]
layer index & 11 & 12 & 13 & 14 & 15 & 16 & 17 & 18 & 19 & 20\\
\cmidrule{1-11}
correlation & 0.011 & 0.184 & -0.014 & 0.001 & 0.012 & 0.024 & 0.011 & -0.001 & -0.071 & -0.017 \\
\cmidrule{1-11}
mean (nume.) & -0.296 & -0.168 & -0.084 & -0.310 & -0.137 & -0.201 & 0.003 & -0.239 & -0.407 & 0.634 \\
mean (single) & -1.117 & -1.134 & -2.417 & -1.437 & -1.043 & -1.977 & -0.596 & 1.898 & -3.203 & 3.358 \\
\cmidrule{1-11}
variance (nume.) & 17.704 & 20.141 & 18.767 & 21.080 & 28.461 & 23.795 & 19.475 & 14.592 & 13.703 & 4.220 \\
variance (single) & 549.51 & 500.61 & 585.75 & 507.40 & 792.44 & 588.06 & 581.66 & 501.08 & 487.62 & 140.33 \\
\bottomrule
\end{tabular}
\end{center}
\vspace{-4mm}
\end{table*}

\subsection{Investigation into the Learned Weight}

In this subsection, we discuss about the relationship between different learned convolutional kernels, and alternatives of the learned parameter position and type to better understand our proposed framework.

\setlength{\tabcolsep}{4pt}
\renewcommand{\arraystretch}{1}
\begin{table*}[t]
\begin{center}
\caption{Numerical analysis of learned parameter type and position. ``norm" and ``conv" indicates the learned parameter type in either the instance normalization layers or convolution layers. The number in the brackets behind the parameter type refer to the position where the learned layer is located, ``1'' for the 1st layer and ``all'' for all the layers.}
\label{table:pami_11}
\begin{tabular}{cl cccccccccc c}
\toprule[0.08em]
\cmidrule{1-13}
metric & method & $L_0$ & WLS & RTV & RGF & WMF & LLF & LLF remap & WLS enhance & Stylization & Abstraction & Average \\
\cmidrule{1-13}
\multirow{9}{*}{\small{PSNR}}
&norm(1) & 31.06 & 34.86 & 33.75 & 32.87 & 33.49 & 29.22 & 32.21 & 32.59 & 27.26 & 24.81 & \textbf{31.21} \\
&norm(7) & 30.86 & 34.75 & 33.54 & 32.97 & 33.39 & 29.24 & 31.99 & 32.79 & 27.07 & 24.81 & \textbf{31.14} \\
&norm(14) & 30.64 & 34.31 & 33.06 & 32.42 & 33.21 & 29.15 & 31.80 & 32.53 & 27.05 & 24.83 & \textbf{30.90} \\
&norm(19) & 28.53 & 31.22 & 29.46 & 29.72 & 30.69 & 27.03 & 30.36 & 29.78 & 25.57 & 24.38 & \textbf{28.67} \\
\cmidrule{2-13}
&conv(19) & 30.57 & 34.17 & 32.45 & 31.84 & 32.96 & 28.80 & 32.07 & 31.95 & 27.46 & 26.30 & \textbf{30.85} \\
&conv channel1(19) & 29.54 & 32.80 & 31.14 & 30.94 & 31.97 & 28.09 & 31.53 & 31.39 & 27.07 & 25.89 & \textbf{30.03} \\
&conv channel2(19) & 24.87 & 26.63 & 24.86 & 26.17 & 27.48 & 23.26 & 24.86 & 25.29 & 21.82 & 24.45 & \textbf{24.96} \\
\cmidrule{2-13}
&norm(all) & 31.71 & 35.51 & 34.31 & 33.10 & 33.96 & 29.56 & 32.45 & 33.11 & 27.32 & 25.21 & \textbf{31.62} \\
&conv(all) & 31.64 & 35.02 & 33.67 & 32.81 & 33.97 & 29.59 & 32.52 & 32.69 & 28.12 & 26.35 & \textbf{31.63} \\
\cmidrule{1-13}
\multirow{9}{*}{\small{SSIM}}
&norm(1)   & 0.949 & 0.971 & 0.959 & 0.954 & 0.948 & 0.966 & 0.981 & 0.982 & 0.923 & 0.819 & \textbf{0.945} \\
&norm(7)   & 0.945 & 0.969 & 0.958 & 0.955 & 0.948 & 0.967 & 0.980 & 0.982 & 0.917 & 0.812 & \textbf{0.943} \\
&norm(14)  & 0.935 & 0.963 & 0.942 & 0.947 & 0.945 & 0.966 & 0.979 & 0.980 & 0.917 & 0.816 & \textbf{0.939} \\
&norm(19)  & 0.871 & 0.890 & 0.809 & 0.881 & 0.876 & 0.947 & 0.970 & 0.965 & 0.889 & 0.796 & \textbf{0.889} \\
\cmidrule{2-13}
&conv(19)  & 0.931 & 0.948 & 0.913 & 0.932 & 0.932 & 0.965 & 0.979 & 0.989 & 0.911 & 0.826 & \textbf{0.932} \\
&conv channel1(19)  & 0.905 & 0.927 & 0.862 & 0.917 & 0.921 & 0.957 & 0.976 & 0.974 & 0.906 & 0.822 & \textbf{0.916} \\
&conv channel2(19)  & 0.741 & 0.727 & 0.629 & 0.700 & 0.737 & 0.844 & 0.899 & 0.914 & 0.848 & 0.741 & \textbf{0.778} \\
\cmidrule{2-13}
&norm(all) & 0.954 & 0.972 & 0.960 & 0.956 & 0.953 & 0.970 & 0.982 & 0.984 & 0.924 & 0.823 & \textbf{0.947} \\
&conv(all) & 0.954 & 0.969 & 0.952 & 0.953 & 0.954 & 0.969 & 0.982 & 0.982 & 0.925 & 0.833 & \textbf{0.947} \\
\bottomrule
\end{tabular}
\end{center}
\vspace{-4mm}
\end{table*}

\subsubsection{Difference of the learned convolution kernels between jointly trained network and solely trained network}\label{sec:kernel_corr}

To analyse the difference of the convolution weights between networks jointly trained on numerous random parameter values (``nume.'') and a single parameter value (``fixed''), we compute their correlation coefficient, individual mean and variance for each layer as shown in Table \ref{table:pami_10}.

As can be seen, the correlation coefficient is almost 0 everywhere, which means there is no linear relationship between the two groups of convolution kernels. The absolute mean and variance of jointly trained network is also significantly smaller than that of the solely trained network. Therefore, in each way, their learned convolution weights are very different from each other, even if the learned smoothing effect is almost the same (PSNR/SSIM: 35.51dB/0.983 (nume.) \textit{vs.} 35.83dB/0.982 (fixed)).

This simple experiment further verifies the huge solution space in the form of learned convolution kernels. Two exactly same results may be represented by very different convolution weights. The linear transformation in our proposed weight learning network actually connects all the different image operators and constrains their learned convolution weights in a limited high dimensional space.

\subsubsection{Study of the learned parameter position and type}

For most image operators presented in the paper, there is only one parameter needed. It might be too complex to use a \textit{weight learning} network that only takes a scalar to manipulate all the convolution layers in the \textit{base} network. To conduct a comprehensive study of the effects of where and what the learned controllable parameters are, we start by three experimental settings, (i). learned instance normalization parameters at different network depth. (ii). learned different types of convolution parameters at a fixed network depth. (iii). learned parameters in either all the convolutional or instance normalization layers.

We compare the performance of different network settings in Table \ref{table:pami_11}. These experiments are made by reproducing the 10 image filters with the base network described in Section \ref{sec:basic}.

\textbf{(i)}. At first, we adjust the position of learned controllable parameters with four discrete network depth. It ranges from the first (1) instance normalization layer to the last (19) layer (the 20th convolutional layer is not followed by instance normalization). From the table, we can see that as the learned parameters become deeper in the network, the performance degrades. It's a reasonable phenomenon, since the deeper the parameters are, the smaller the network capacity to differentiate various image operators is, and thus it's harder to incorporate all the operators in the network.

We can also observe that the average error (PSNR) over all the 10 image operators only decreases by about 2.5dB from the first layer to the last layer, however it significantly improves the running speed when varying between different image operators in the network.

\textbf{(ii)}. To explore the influence of other types of parameters in the network to separate these image operators, we replace the controllable parameters in the instance normalization layer with the ones in the convolutional layer.

Each convolutional layer contains a $M$$\times$$N$$\times$$k$$\times$$k$ weight tensor and a $M$-length bias vector. $M$ is the number of output feature channels, $N$ is the number of input feature channels, $k$ is the kernel size. $M$ equals to $N$ for the intermediate convolutional layers in our baseline network. Since the convolutional layer is followed by an instance normalization layer, the added bias in the convolutional layer will be balanced out by the following normalization operation and becomes useless. Therefore we experiment only by varying the parameters in the convolutional weights.

\setlength{\tabcolsep}{5pt}
\begin{table*}[t]
\begin{center}

\caption{ Quantitative evaluation of a few variants of the proposed network trained on the WLS filter \cite{farbman2008edge}. We experiment separately by training only the base network on a fixed single parameter value (``single (base)''), extending the weight learning network to two or more fully connected layers trained on numerous random parameter values with (``nume. (fc2R)'') and without (``nume. (fc2)'') ReLU layers inbetween.}
\label{table:pami_13}

\begin{tabular}{c cccccccccc}
\toprule[0.08em]
$\lambda$ & \multicolumn{2}{ c }{{single}} & \multicolumn{2}{ c }{single (base)} & \multicolumn{2}{ c }{{nume.}} & \multicolumn{2}{ c }{{nume. ({fc2R})}} & \multicolumn{2}{ c }{{nume. (fc2)}}\\
\cmidrule{1-11}
& {PSNR} & {SSIM} & {PSNR} & {SSIM} & {PSNR} & {SSIM} & {PSNR} & {SSIM} & {PSNR} & {SSIM} \\
\cmidrule{1-11}
0.100 & 44.00&0.994 & 44.16&0.994 & 42.12&0.993 &42.02&0.992 &42.36&0.993 \\
0.215 & 43.14&0.993 & 43.08&0.993 & 42.64&0.993 &42.23&0.993 &42.43&0.993 \\
1.000 & 41.93&0.992 & 40.61&0.991 & 41.63&0.991 &40.25&0.991 &40.28&0.991 \\
4.641 & 39.42&0.987 & 38.01&0.988 & 39.64&0.989 &37.31&0.988 &37.35&0.988 \\
10.00 & 39.13&0.986 & 36.83&0.986 & 38.51&0.987 &36.00&0.985 &35.94&0.986 \\
\cmidrule{1-11}
average & 41.52&0.990 & 40.54&0.990 & 40.91&0.990 &39.56&0.990 &39.67&0.990 \\
\bottomrule
\end{tabular}
\end{center}
\vspace{-4mm}
\end{table*} 
\setlength{\tabcolsep}{3pt}
\begin{table*}[htp]
\begin{center}
\caption{Quantitative evaluation of our proposed framework trained only with a set of fixed parameter values (``various (fixed)'') on the $L_0$ smoother \cite{xu2011image}. The parameters used for training our framework are taken from the 5 non-boldface parameters between [0.002, 0.2] in the table. The extra 4 parameters with boldface are only used in the test stage. The absolute difference between the network trained on a single parameter (``single'') and various fixed parameters (``various (fixed)'') is displayed in the bottom. }
\label{table:pami_12}
\begin{tabular}{lc ccccccccccc}
\toprule[0.08em]
& & 0.0020 & \textbf{0.0025} & \textbf{0.0033} & 0.0043 & 0.0200 & 0.0928 & \textbf{0.1200} & \textbf{0.1600} & 0.2000  & average\\
\cmidrule{1-12}
\multirow{2}{*}{single}
& PSNR & 40.69 & 40.19 & 39.77 & 38.96 & 36.07 & 33.08 & 31.78 & 31.13 & 31.75 & 35.94 \\
& SSIM & 0.989 & 0.987 & 0.986 & 0.986 & 0.982 & 0.977 & 0.973 & 0.972 & 0.973 & 0.981 \\
\cmidrule{1-12}
\multirow{2}{*}{various (fixed)}
& PSNR & 39.61 & 39.33 & 38.95 & 38.51 & 35.37 & 31.80 & 31.40 & 30.69 & 30.54 & 35.13 \\
& SSIM & 0.988 & 0.987 & 0.986 & 0.986 & 0.979 & 0.972 & 0.972 & 0.971 & 0.970 & 0.979 \\
\cmidrule{1-12}
\multirow{2}{*}{difference}
& PSNR & 1.08  & \textbf{0.86}  & \textbf{0.82}  & 0.45  & 0.7   & 1.28  & \textbf{0.38}  & \textbf{0.44}  & 1.21  & 0.81  \\
& SSIM & 0.001 & \textbf{0}     & \textbf{0}     & 0     & 0.003 & 0.005 & \textbf{0.001} & \textbf{0.001} & 0.003 & 0.002 \\
\bottomrule
\end{tabular}
\end{center}
\end{table*}

We study with three types of learned convolutional parameters, which are separately: ``conv'', all the weight kernels ($M$$\times$$N$$\times$$k$$\times$$k$); ``conv channel1'', one slice of weight kernels ($M$$\times$$1$$\times$$k$$\times$$k$); ``conv channel2'', another slice of weight kernels ($1$$\times$$N$$\times$$k$$\times$$k$). The learned convolutional layer shares the same network depth as the last instance normalization layer.

As can be seen from the table, learning all the weight kernels in the convolutional layer (conv) achieves the highest performance, which is very reasonable. While learning all the convolutional kernels that are only related to one single input channel (conv channel1), the performance degrades by about 0.8dB. Learning all the parameters that are related to one single output channel (conv channel2) obtains the worst performance. Note even if ``conv channel1'' and ``conv channel2'' share the same number of controllable parameters, all the learned parameters in ``conv channel2'' only result in adjustment of a single channel of output feature map, which influence the following layers much less than ``conv channel1'' and hence achieve worse results.

Interestingly, by learning the convolutional parameters (conv and conv channel1) achieves better results than learning instance normalization parameters at the same network depth. This may be because the convolutional layer is ahead of the instance normalization, and the learned convolutional parameters (even $M$$\times$$k$$\times$$k$ for ``conv channel1'', $k$ equals to 3 in our case) is much more than the normalization's ($M$$\times$$2$).

\textbf{(iii)}. To further explore the difference between learning convolution and normalization layers, we try to learn the parameters in either all the convolutional layers or all the instance normalization layers in the network, which should theoretically achieves the best performance of incorporating all the image operators.

As shown in the ``conv(all)'' and ``norm(all)'' rows, surprisingly they achieve almost the same performance on the either PSNR or SSIM error metric. This means it already touches the top performance limit of incorporating the 10 image operators with this neural network.

Note learning all the normalization parameters (norm(all)) only obtains a little improvement (0.41dB) than learning the one in the first layer (norm(1)), from which we can see it's more important which position the controllable parameters are instead of how many layers the parameters are located in.

\begin{figure*}[t]
	\includegraphics[width=0.9\linewidth]{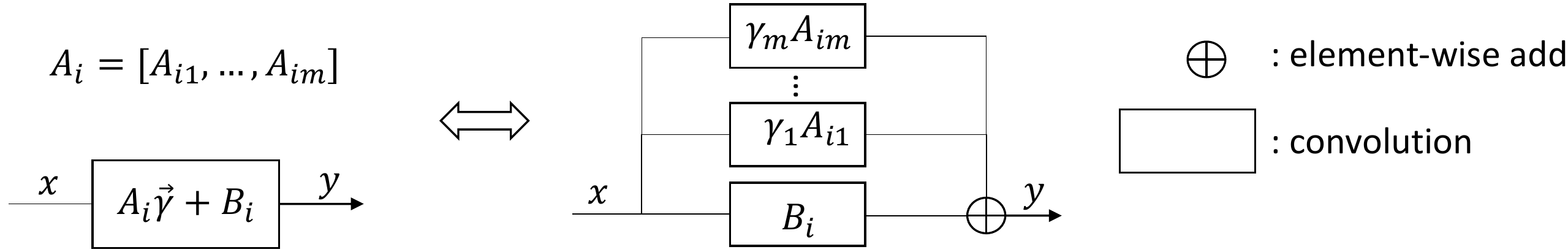}
	\caption{Equivalent analysis of the connection between the \textit{base} network $\mathcal{N}_{base}$ and the \textit{weight learning} network $\mathcal{N}_{weight}$. One convolution layer whose weights are learnt by the fc layer is exactly equivalent to a multi-path convolution blocks.}
	\label{figure:analysis}
\end{figure*}

\subsection{Interpretation of the Weight Learning Network}
To help understand the connection between the \textit{base} network $\mathcal{N}_{base}$ and the \textit{weight learning} network $\mathcal{N}_{weight}$, we decompose the parameter vector $\overrightarrow{\gamma}$ and the weight matrix $A_i$ into independent elements $\gamma_1,...,\gamma_m$ and $A_{i0}, ...,A_{im}$ respectively, then:

\begin{equation}
\begin{aligned}
(A_i\overrightarrow{\gamma} + B_i)\otimes x = \sum_{k=1}^m \gamma_k A_{ik} \otimes x + B_i\otimes x
\end{aligned}
\label{fg:fc_analysis}
\end{equation}

where $\otimes$ denotes convolution operation, and $m$ is the dimension of $\overrightarrow{\gamma}$. In other words, the one convolution layer, whose weights are learned with one single fc layer, is exactly equivalent to a multi-path convolution block as shown in Figure \ref{figure:analysis}. Learning the weight and bias of the single fc layer is equivalent to learning the common basic convolution kernels $B_i, A_{i1}, A_{i2},...,A_{im}$ in the convolution block.

\subsection{Analysis of More Variants of Our Proposed Network}

In this subsection, we experiment with a few variants of our network to justify its effectiveness.

\subsubsection{Training the \textit{base} network only} Since training a fully convolutional network alone has been employed frequently by previous image processing papers \cite{liu2016learning,fan2017generic,chen2017fast}, which presents a strong baseline, we experiment with this alternative (``nume. (base) '') which only leverages the \textit{base} network by training it on one specific parameter configuration. As show in Table \ref{table:pami_13}, our proposed framework (``single'') achieves even better performance than the baseline under the PSNR error metric.

\subsubsection{Training with more fc layers} We also try a deeper \textit{weight learning} network with more fully connected layers. Here we simply add one more fully connected layer to the default \textit{weight learning} network, and demonstrate the performance of its two variants with (``nume. (fc2R)'') and without (``nume. (fc2)'') ReLU between these two layers respectively. As shown in Table \ref{table:pami_13}, they all achieve comparable performance with that of single layer (``nume.'') used in our paper. The potential reason for this phenomenon is that this one-layer \textit{weight learning} network is sufficient for adaptively learning various parameter settings, while adding more weights/complexity to the network does not contribute to the performance much.

\subsection{Interpolation Ability of the Proposed Framework on Unseen Input Parameters}

Since the \textit{weight learning} network contains a single fully connected layer with no non-linear activation layers, the predicted convolution weights should be a linear transformation of the input parameters. Given such a fact, any convolutional kernels of a specific parameter should be the linear interpolation of the other two. Hence we are curious about the interpolation ability of the proposed framework.

To verify such a property, we train the network using only a few fixed parameter values, which corresponds to the 5 non-boldface parameter values in Table \ref{table:pami_12}. But in the test stage, we use another 4 parameter values (boldface in Table \ref{table:pami_12}) that have not been seen by the network in the training stage but are between the lower and upper bound of its parameter range. As shown in Table \ref{table:pami_12}, we can see that the network performs similarly to the one trained with only one parameter value, but more importantly for the interpolated boldface parameter values that the network does not recognize, it also surprisingly achieves very comparable results.

This means that a few parameter values are already sufficient for learning a good linear transformation in the \textit{weight learning} network from input parameters to convolution weights. However, as in a real scenario, the number of such fixed training parameters is usually difficult to decide, due to many different parameter ranges of image operators. As a result, we choose to sample random parameter values instead of only a few of them for training in our paper.

\section{Conclusion}
In this paper, we propose the first decouple learning framework for parameterized image operators, where the weights of the task-oriented \textit{base} network $\mathcal{N}_{base}$ are decoupled from the network structure and directly learned by another \textit{weight learning} network $\mathcal{N}_{weight}$. These two networks can be easily end-to-end trained, and $\mathcal{N}_{weight}$ dynamically adjusts the weights of $\mathcal{N}_{base}$ for different parameters $\overrightarrow{\gamma}$ during the runtime. We show that the proposed framework can be applied to different parameterized image operators, such as image smoothing, denoising and super resolution, while obtaining comparable performance as the network trained for one specific parameter configuration. It also has the potential to jointly learn multiple different parameterized image operators within one single network. For real user scenarios, we further extend our framework to enable cheap parameter tuning, which obtains superior performance over state-of-the-art methods by a large margin. To better understand the working principle, we also provide some valuable analysis and discussions, which may inspire more promising research in this direction. More theoretical analysis is worthy of further exploration in the future.

{
\bibliographystyle{IEEEtran}
\bibliography{egbib}
}

\vspace{-1cm}
\begin{IEEEbiography}[{\includegraphics[width=1in,height=1.25in,clip,keepaspectratio]{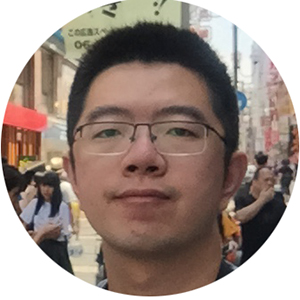}}]{Qingnan Fan} is a Postdoctoral Scholar in the Computer Science Department of Stanford University. He received his PhD degree from Shandong University in 2019. His research interests mainly include image/video processing and 3D vision.
\end{IEEEbiography}
\vspace{-1cm}
\begin{IEEEbiography}[{\includegraphics[width=1in,height=1.25in,clip,keepaspectratio]{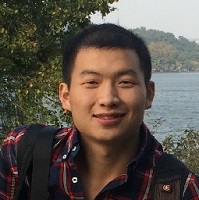}}]{Dongdong Chen} is a Ph.D. student from the University of Science and Technology of China. He is also a joint PhD between his university and Microsoft Asia. His research interests mainly include style transfer, image generation, low-level image processing and object detection.
\end{IEEEbiography}
\vspace{-1cm}
\begin{IEEEbiography}[{\includegraphics[width=1in,height=1.25in,clip,keepaspectratio]{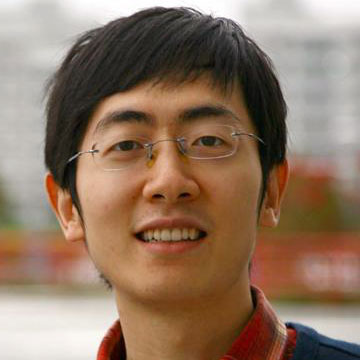}}]{Lu Yuan} received his PhD degree from the Department of Computer Science and Engineering at the Hong Kong University of Science and Technology in 2009. Now he is a Senior Research Manager in Microsoft Redmond. His research interests include computer vision, applied machine learning and computational photography.
\end{IEEEbiography}
\vspace{-1cm}
\begin{IEEEbiography}[{\includegraphics[width=1in,height=1.25in,clip,keepaspectratio]{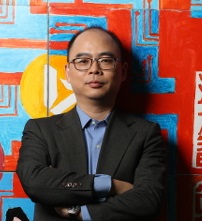}}]{Gang Hua} is the Vice President and Chief Scientist of Wormpex AI Research. He was an Associate Professor of Computer Science in Stevens Institute of Technology between 2011 and 2015, while holding an Academic Advisor position at IBM T. J. Watson Research Center. He has published more than 130 peer reviewed papers in top conferences such as CVPR/ICCV/ECCV, and top journals such as T-PAMI and IJCV. To date He holds 18 issued U.S Patents and also has 14 more U.S. Patents Pending. He is an IAPR Fellow, an ACM Distinguished Scientist, and a Senior Member of IEEE. His research focuses on artificial intelligence, computer vision, pattern recognition, machine learning, and robotics, with primary applications in the cloud and mobile intelligence domain.
\end{IEEEbiography}
\vspace{-1cm}
\begin{IEEEbiography}[{\includegraphics[width=1in,height=1.25in,clip,keepaspectratio]{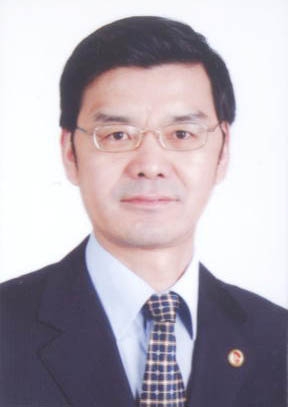}}]{Nenghai Yu} is a full Professor at University of Science and Technology of China. He is also the director of Information Processing Center of USTC,deputy director of academic committee of School of Information Science and Technology. He was a visiting scholar in Institute of Production Technology, Faculty of Engineering, University of Tokyo, in 1999 and did cooperative research as the senior visiting scholar in Dept. of Electrical Engineering, Columbia University, from Apr. to Oct. 2008. His research focuses on image processing and video analysis, multimedia communication, media content security, Internet information retrieval, data mining and content filtering, network communication and security.
\end{IEEEbiography}
\vspace{-1cm}
\begin{IEEEbiography}[{\includegraphics[width=1in,height=1.25in,clip,keepaspectratio]{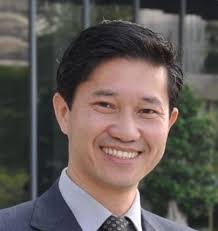}}]{Baoquan Chen} is a Professor of Peking University, where he is the Executive Director of the Center on Frontiers of Computing Studies. Prior to the current post, he was the Dean of School of Computer Science and Technology at Shandong University, and the founding director of the Visual Computing Research Center, Shenzhen Institute of Advanced Technology (SIAT), Chinese Academy of Sciences (2008-2013), and a faculty member at Computer Science and Engineering at the University of Minnesota at Twin Cities (2000-2008). His research interests generally lie in computer graphics, visualization, and human-computer interaction, focusing specifically on large-scale city modeling, simulation and visualization.
\end{IEEEbiography}

\end{document}